\documentclass{article}

\usepackage{arxiv}
\usepackage[utf8]{inputenc}
\usepackage[T1]{fontenc}
\usepackage{textcomp}

\usepackage[utf8]{inputenc} % allow utf-8 input
\usepackage[T1]{fontenc}    % use 8-bit T1 fonts
\usepackage{hyperref}       % hyperlinks
\usepackage{url}            % simple URL typesetting
\usepackage{booktabs}       % professional-quality tables
\usepackage{amsfonts}       % blackboard math symbols
\usepackage{nicefrac}       % compact symbols for 1/2, etc.
\usepackage{microtype}      % microtypography
\usepackage{lipsum}
\usepackage{graphicx}
\usepackage{tabularx}
\usepackage{authblk}
\usepackage{amsmath} 
\usepackage{booktabs}
\usepackage{cite}

\usepackage{algorithmicx}
\usepackage{algpseudocode}
\usepackage{algorithm}
\usepackage{xcolor}
\definecolor{DeepGreen}{RGB}{0,110,0} 
\newcommand{\best}[1]{\textcolor{red}{#1}}
\newcommand{\second}[1]{\textcolor{blue}{#1}}
\newcommand{\third}[1]{\textcolor{DeepGreen}{#1}}

\graphicspath{ {./images/} }

\title{Prior-Guided DETR for Ultrasound Nodule Detection}

\author[1]{Jingjing Wang}
\author[1]{Zhuo Xiao}
\author[1]{Xinning Yao}
\author[1,2,*]{Bo Liu}
\author[3,*]{Lijuan Niu}
\author[1,4,5,*]{Xiangzhi Bai}
\author[1,2]{Fugen Zhou}

\affil[1]{Image Processing Center, Beihang University, Beijing 102206, China}
\affil[2]{State Key Laboratory of High-Efficiency Reusable Aerospace Transportation Technology, Beijing 102206, China}
\affil[3]{Department of Ultrasound, National Cancer Center/National Clinical Research Center for Cancer/Cancer Hospital, Chinese Academy of Medical Sciences and Peking Union Medical College, Beijing 100021, China}
\affil[4]{State Key Laboratory of Virtual Reality Technology and Systems, Ministry of Education, Beihang University, Beijing 102206, China}
\affil[5]{the Key Laboratory of Spacecraft Design Optimization and Dynamic Simulation Technology, Ministry of Education, Beihang University, Beijing 102206, China}
\affil[ ]{\textbf{\texttt{bo.liu@buaa.edu.cn, niulijuan@cicams.ac.cn, jackybxz@buaa.edu.cn}}}

\begin{document}
\maketitle
\begin{abstract}
Accurate detection of ultrasound nodules is essential for the early diagnosis and treatment of thyroid and breast cancers. However, this task remains challenging due to irregular nodule shapes, indistinct boundaries, substantial scale variations, and the presence of speckle noise that degrades structural visibility. To address these challenges, we propose a prior-guided DETR framework specifically designed for ultrasound nodule detection. Instead of relying on purely data-driven feature learning, the proposed framework progressively incorporates different prior knowledge at multiple stages of the network. First, a Spatially-adaptive Deformable FFN with Prior Regularization (SDFPR) is embedded into the CNN backbone to inject geometric priors into deformable sampling, stabilizing feature extraction for irregular and blurred nodules. Second, a Multi-scale Spatial-Frequency Feature Mixer (MSFFM) is designed to extract multi-scale structural priors, where spatial-domain processing emphasizes contour continuity and boundary cues, while frequency-domain modeling captures global morphology and suppresses speckle noise. Furthermore, a Dense Feature Interaction (DFI) mechanism propagates and exploits these prior-modulated features across all encoder layers, enabling the decoder to enhance query refinement under consistent geometric and structural guidance. Experiments conducted on two clinically collected thyroid ultrasound datasets (Thyroid I and Thyroid II) and two public benchmarks (TN3K and BUSI) for thyroid and breast nodules demonstrate that the proposed method achieves superior accuracy compared with 18 detection methods, particularly in detecting morphologically complex nodules. These results highlight the effectiveness and generalizability of progressively integrating prior knowledge for robust ultrasound nodule detection. The source code is publicly available at https://github.com/wjj1wjj/Ultrasound-DETR.
\end{abstract}

\section{Introduction}
Thyroid and breast cancers rank among the most prevalent malignancies globally \cite{15liu2019automated} \cite{64Lin2022}. Accurate and timely detection of nodules is critical for early diagnosis, clinical decision-making, and improving patient outcomes \cite{4nabhan2021thyroid} \cite{63Cao2020}. Ultrasound (US) is widely used for initial screening due to its non-invasiveness, low cost, and real-time imaging capability \cite{62Chang2020}. However, the diagnostic interpretation of ultrasound images relies heavily on manual assessment of morphological characteristics. This process is inherently susceptible to inter-observer variability and is heavily reliant on the subjective experience of the radiologist. This study addresses both thyroid and breast nodules, as they present similar challenges for automated systems. Clinically, nodules in both superficial organs are assessed using highly analogous sonographic features (e.g., hypoechogenicity, indistinct margins, irregular shapes, and complex internal textures) under similar high-frequency imaging conditions. This similarity is reflected in the design of TI-RADS \cite{7grant2015thyroid} and BI-RADS \cite{8d2018breast}. Therefore, developing a powerful automated detection method capable of addressing these shared challenges is crucial for standardizing diagnosis and reducing missed detections across both organs.

Traditional machine learning methods relied on handcraf3lin2017focalted features \cite{9Zhang2019} \cite{10Ouyang2019}, which achieved limited success due to their restricted representational capacity and dependence on manual feature design. These approaches lacked robustness to irregular morphology, subtle echogenic variations, and inter-patient heterogeneity, motivating the adoption of deep learning. Convolutional neural networks (CNNs) and their variants have achieved remarkable success in medical image analysis, including ultrasound nodule detection. CNN-based detectors such as Faster R-CNN \cite{12ren2016faster}, RetinaNet \cite{13lin2017focal}, and the YOLO series \cite{14redmon2016you} have been widely adapted for ultrasound images \cite{15liu2019automated} \cite{56wang2023thinking}, yielding notable improvements. Nevertheless, these approaches still rely on predefined anchor boxes and post-processing strategies like non-maximum suppression (NMS), which may limit their adaptability to diverse imaging conditions and nodule characteristics. Transformer, leveraging its self-attention mechanism,  excels at modeling the long-range dependencies that CNNs struggle with, but often struggles to preserve fine-grained spatial details. This inherent trade-off between CNN's strong local representation and Transformer's global context modeling capability has motivated the development of hybrid approaches.
	
The Detection Transformer (DETR) \cite{19carion2020end} introduced a new end-to-end detection paradigm by formulating the task as a direct set prediction problem, eliminating the need for anchors and NMS. DETR employs a CNN backbone for feature extraction while leveraging the Transformer's self-attention to capture long-range dependencies. This paradigm shift has inspired extensive research in both natural and medical imaging. Nonetheless, vanilla DETR still suffers from slow convergence and limited performance on small or irregularly shaped nodules \cite{20meng2021conditional}, which are common in ultrasound images. Although numerous variants—such as Deformable DETR \cite{21zhu2020deformable} and DN-DETR \cite{22li2022dn}—have been proposed to improve convergence speed and strengthen multi-scale detection, several ultrasound-specific challenges remain unresolved:

\begin{itemize}
    \item Lack of geometric priors under irregular morphology: Ultrasound nodules often present irregular shapes and blurred boundaries exacerbated by speckle noise and acoustic artifacts. Without explicit geometric priors, detectors with fixed receptive fields or unconstrained deformable sampling struggle to capture nodule morphology \cite{24xie2016ultrasonography}.
    
    \item Insufficient structural priors for physics-driven multi-scale heterogeneity: Nodule appearance varies across scales and interacts with ultrasound physics in a frequency-dependent manner, where speckle corrupts high-frequency details and acoustic shadowing obscures low-frequency morphology \cite{26Das2024}.  Without structural priors that jointly model contour continuity and global morphology, existing methods have difficulty separating anatomical structure from physics-induced artifacts.
    
    \item Underutilization of prior-modulated multi-level features: DETR-like models mostly use only the final encoder layer for decoding, underutilizing the rich, prior-modulated, multi-level semantic information from earlier layers \cite{27nan2025mi}.

\end{itemize}

\begin{figure*}[htbp]
    \centering
    \includegraphics[width=\textwidth]{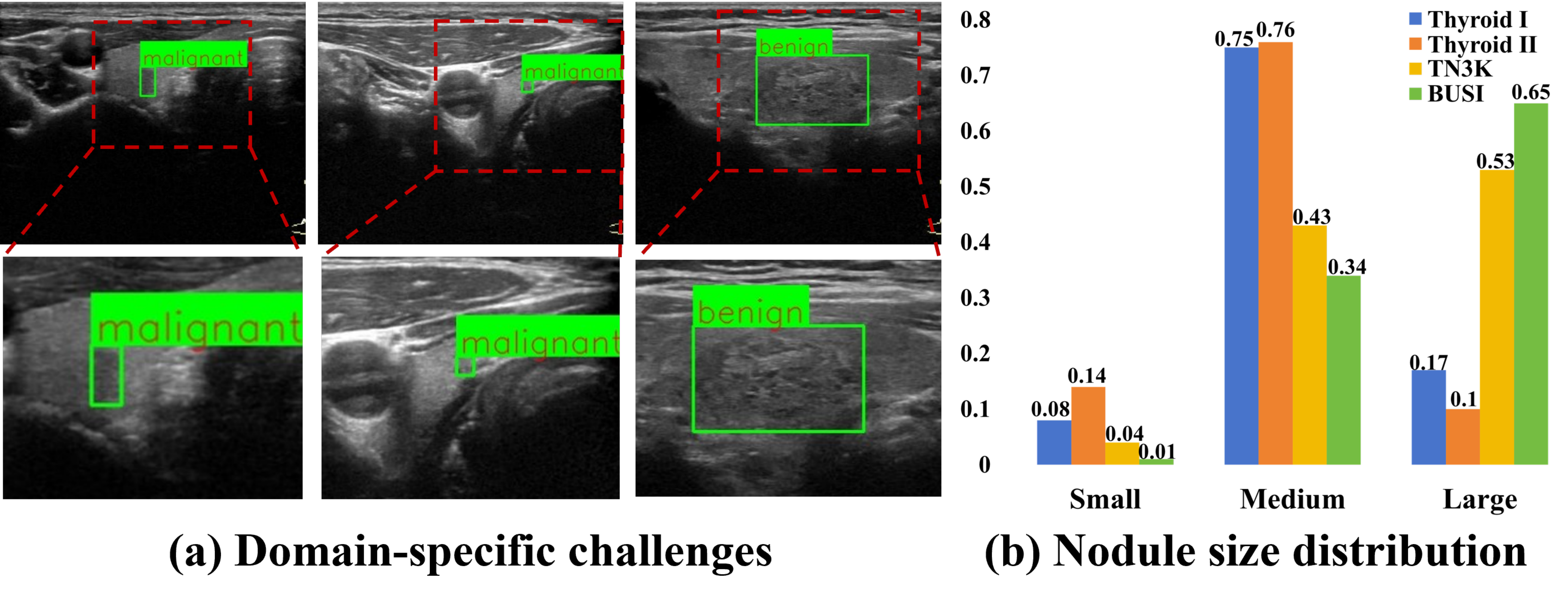} 
    \caption{Typical challenges in ultrasound nodule detection. Irregular morphology, blurred boundaries, and strong speckle noise jointly degrade structural visibility, while pronounced scale variation further complicates robust localization. These characteristics reveal a fundamental mismatch between ultrasound image formation physics and purely data-driven detection models, motivating the explicit incorporation of geometric and structural priors into the detection framework.} 
    \label{fig:fig1} 
\end{figure*}
	
These challenges are summarized in Fig. 1, revealing a fundamental mismatch between the physics-driven formation of ultrasound images and the implicit assumptions of purely data-driven detection models.
	
To address these domain-specific challenges, we propose a prior-guided DETR framework tailored for ultrasound nodule detection. The core idea is to progressively introduce and exploit different forms of prior knowledge throughout the detection pipeline, rather than relying solely on data-driven attention mechanisms. First, we design a SDFPR module that augments DCNv4 \cite{28xiong2024efficient} with geometric priors to improve feature extraction from irregular morphologies and blurred boundaries. Second, we develop a MSFFM to provide structural priors, which combines a spatial branch for fine structural details with a frequency-domain branch that enhances global morphology and suppresses speckle-dominated high-frequency noise, thereby strengthening multi-scale robustness. Finally, inspired by DenseNet \cite{29huang2017densely}, we propose a DFI mechanism that propagates and exploits these prior-modulated features across multiple encoder layers. Instead of relying solely on the final encoder output, DFI ensures that geometric and structural priors learned at different depths consistently guide decoder query refinement. Through this prior-modulated design, the proposed framework achieves superior robustness and accuracy for morphologically complex ultrasound nodules.
	
The main contributions of this work can be summarized as follows:

\begin{itemize}
    \item We design a SDFPR module by introducing geometric priors (aspect ratio prior and width prior) into deformable convolution to stabilize the backbone's ability to capture irregular and blurred nodule morphologies.
    
    \item We propose a MSFFM to provide structural priors by jointly modeling spatial contour information and frequency-domain morphology to achieve multi-scale feature fusion, making it uniquely suited for the scale variation and artifact-dependent nature of ultrasound data.
    
    \item We develop a DFI mechanism that propagates and exploits prior-modulated features across all encoder layers, enabling decoder queries to be refined under consistent geometric and structural guidance rather than relying only on the final encoder representation, which maximizes encoder-decoder feature interaction and multi-level reasoning.

    \item We validate our model on four ultrasound datasets, including two clinically collected thyroid datasets for internal validation and two public benchmarks for external and cross-organ generalization, demonstrating the robustness and generalizability of the proposed prior-driven framework.

\end{itemize}

\section{Related Works}
\subsection{CNN-based detection in Ultrasound Imaging}
Liu et al. \cite{15liu2019automated} proposed a clinical-knowledge-guided CNN to automatically detect and classify thyroid nodules in ultrasound images. Wu et al. \cite{39Wu2021} proposed a cache tracking post-processing method that exploits interframe contextual information to propagate detection results to neighboring video frames, thereby improving the accuracy of thyroid nodule detection. Gao et al. \cite{40gao2021detection} developed a semi-supervised model for accurately identifying breast nodules. While CNN-based approaches have shown promising results, their reliance on local receptive fields limits their ability to capture long-range dependencies, which are particularly important when nodules exhibit blurred boundaries or irregular morphology.

\subsection{Transformer-based detection in Ultrasound Imaging}
Gelan et al. \cite{41ayana2022buvitnet} integrated multi-task learning with ViT to perform ultrasound-based breast nodule detection. Meshrif et al. \cite{42alruily2024enhancing} adopted progressive fine-tuning to enable the model to gradually adapt to subtle differences in breast tissue classification, thereby enhancing the detection performance of breast cancer. Feres et al. \cite{43jerbi2023automatic} leveraged a combination of ViT and Generative Adversarial Network (GAN) to achieve automatic classification of ultrasound Thyroid Images. These methods highlight the strength of Transformer in capturing global contextual information and long-range dependencies. However, they often lack strong local perception, which is crucial for accurately modeling fine structural details of nodules.

\subsection{DETR-based detection in Ultrasound Imaging}
Zhou et al. \cite{31zhou2024thyroid} proposed Thyroid-DETR, which applied DETR to ultrasound thyroid nodule detection, verifying the feasibility of DETR in this domain. More recently, DETR variants \cite{44ramezani2024lung} have been applied to medical imaging tasks such as lung and thyroid nodule detection. These models eliminate the need for handcrafted anchors and NMS, while leveraging global self-attention. However, most existing DETR-based methods directly adopt generic attention or multi-scale designs originally developed for natural images, without explicitly incorporating ultrasound-specific prior knowledge. Specifically, the deformable attention used in standard Deformable DETR \cite{21zhu2020deformable} lacks geometric regularization, often leading to unstable and inconsistent sampling in the presence of speckle noise and anisotropic acoustic artifacts. Moreover, conventional pyramid or multi-scale fusion strategies are inherently size-oriented and overlook the frequency-dependent boundary degradations, where speckle noise disrupts high-frequency detail and shadowing degrades low-frequency morphology. Finally, most DETR variants rely solely on the final encoder layer for decoding, resulting in underutilization of prior-enhanced features embedded at earlier stages.

In summary, existing CNN-, Transformer-, and DETR-based methods have made important progress in ultrasound nodule detection, but they largely rely on implicit, data-driven feature learning and lack explicit modeling of domain-specific priors. In particular, geometric regularities under irregular morphology, structural cues across spatial and frequency domains, and the effective utilization of prior-enhanced multi-level features remain insufficiently explored. These limitations motivate a unified prior-guided detection framework, which is detailed in the following section.
 
\section{Methods}
\subsection{Overall Framework}
\begin{figure*}[htbp]
    \centering
    \includegraphics[width=\textwidth]{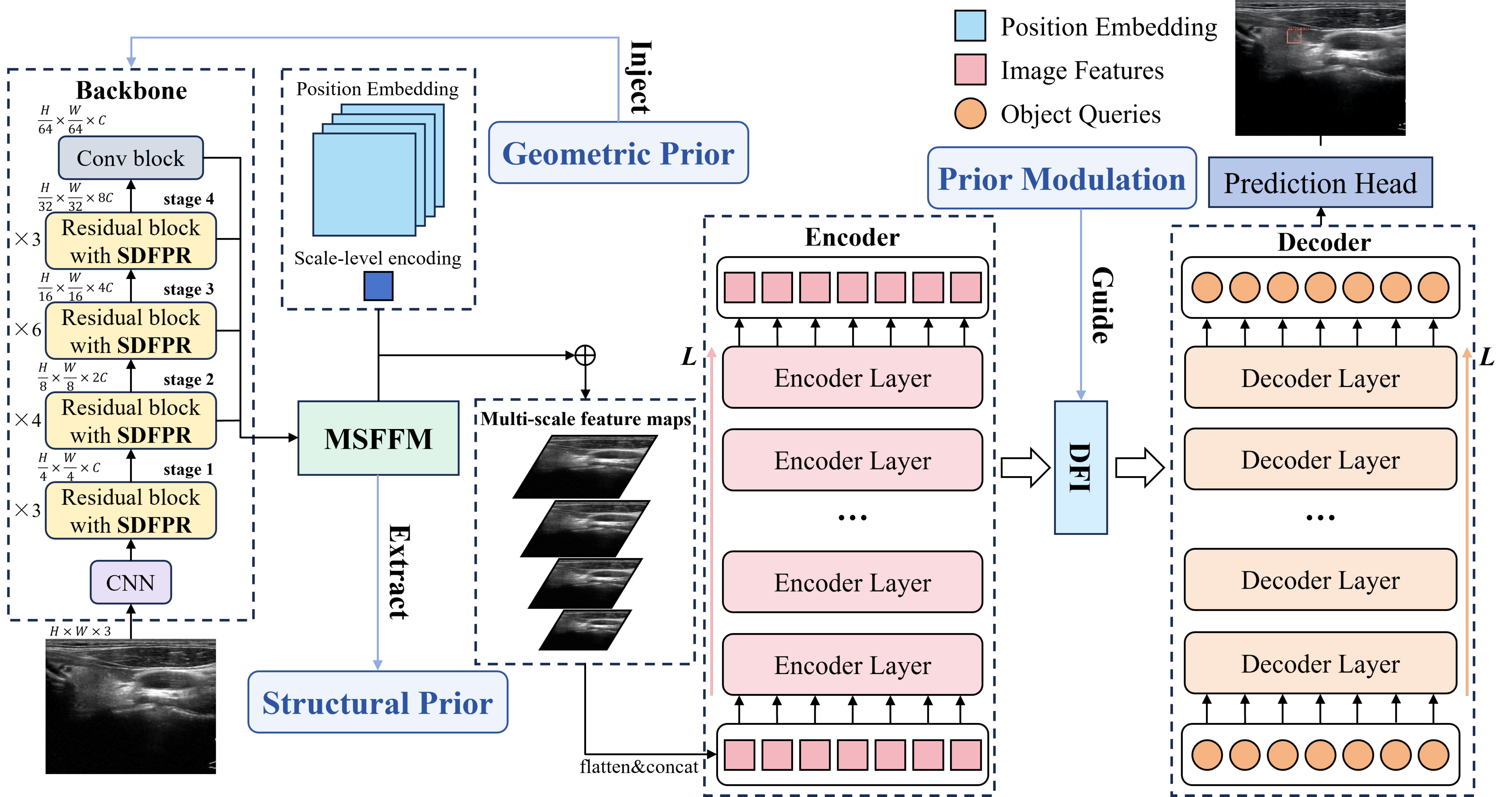} 
    \caption{Overview of the proposed prior-guided DETR. To address the mismatch between ultrasound physics and implicit feature learning, our framework progressively injects domain knowledge at three hierarchical stages: 1) SDFPR embeds Geometric Prior into the backbone to stabilize deformable sampling; 2) MSFFM extracts Structural Prior by synergizing spatial boundary cues with frequency-domain morphology; and 3) DFI propagates these prior-modulated features to the decoder via dense interaction. This unified paradigm ensures robust detection against irregular morphology, speckle noise and multi-scale variation.} 
    \label{fig:fig2} 
\end{figure*}

As illustrated in Fig. 2, we propose a prior-guided DETR framework for ultrasound nodule detection, which progressively integrates different forms of prior knowledge into the detection pipeline. The overall architecture follows an encoder-decoder paradigm, where prior information is introduced, refined, and exploited at multiple stages rather than being implicitly learned in a purely data-driven method.

First, the ResNet50 backbone extracts hierarchical representations from the input ultrasound images. To enhance feature discrimination for irregularly shaped nodules, we embed the SDFPR module into each residual block. Unlike conventional convolutions, SDFPR leverages the dynamic sampling capability of DCNv4 \cite{28xiong2024efficient}, while incorporating aspect ratio and width priors to regularize offset learning, thereby injecting geometric priors into deformable convolution. This design stabilizes sampling under blurred boundaries and irregular morphology, resulting in more robust and discriminative feature maps.

Building on these geometry-aware features, a MSFFM is designed to extract structural priors, including contour continuity and boundary cues in the spatial domain as well as global morphology in the frequency domain. The resulting prior-enhanced representations are further processed by a Transformer encoder, and a DFI mechanism aggregates features across encoder layers so that decoder query refinement is consistently modulated by previously introduced geometric and structural priors.
	
Finally, the decoder receives the refined features and iteratively updates object queries. The output of the final decoder layer is fed into prediction heads to generate bounding boxes and categories of detected nodules.

\subsection{Geometric Prior Injection via Spatially-adaptive Deformable FFN with Prior Regularization}
Ultrasound nodules often exhibit irregular shapes and blurred boundaries arising from anisotropic acoustic propagation and operator-dependent probe orientation, making accurate feature sampling challenging \cite{61Lu2022}. Although DCNv4 \cite{28xiong2024efficient} can adapt to geometric variations by learning offsets dynamically, its unconstrained regression of deformation fields forces the model to search for optimal sampling positions in an unbounded space. The sampling location for each kernel point is learned as $p=p_0+p_k+\Delta p_k$, where  $p_0$ is the center and $\Delta p_k$ is predicted directly from the input feature map. Although this design provides high flexibility, the lack of constraints makes the offset learning process unstable, as it must learn optimal sampling locations from an unbounded search space without any guidance.

\begin{figure*}[htbp]
    \centering
    \includegraphics[width=\textwidth]{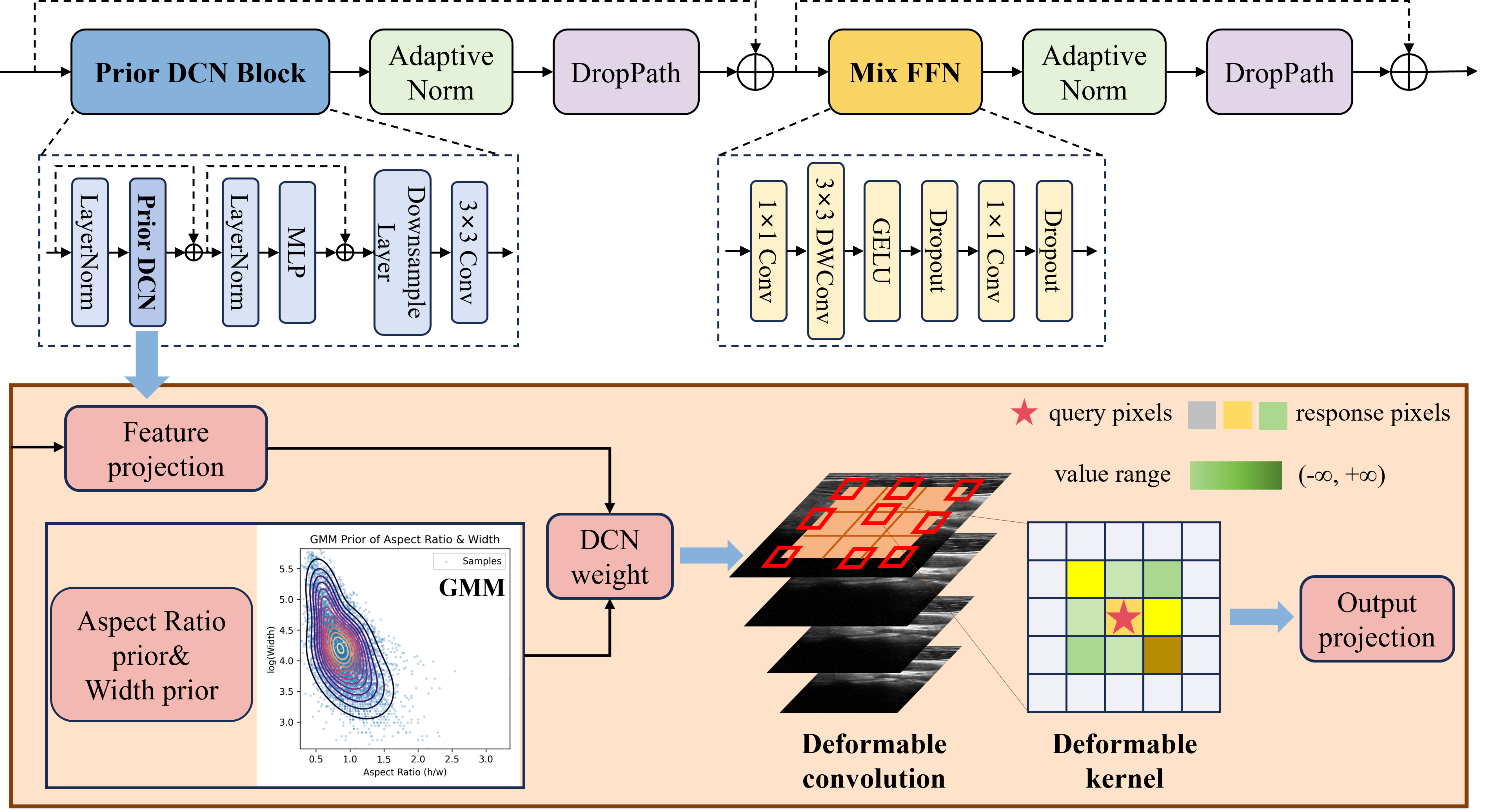} 
    \caption{Structure of the Spatially-adaptive Deformable FFN with Prior Regularization (SDFPR). Conventional deformable convolutions often suffer from unstable offset regression. By embedding geometric priors (aspect ratio and width) learned from clinical data into deformable convolution, SDFPR regularizes offset learning and stabilizes geometric modeling for nodules with irregular shapes and blurred boundaries.} 
    \label{fig:fig3} 
\end{figure*}
	
Existing prior based designs typically introduce hand-crafted geometric assumptions or regularization terms to restrict the deformation field \cite{68Wang2024}. However, such priors are often not derived from population level statistics and may not generalize across organs or imaging conditions. In contrast, the proposed Prior DCN regularizes offset learning using statistical priors that are explicitly learned from clinical data. Instead of regressing unconstrained offsets, Prior DCN modulates the sampling pattern using two geometric priors: an aspect ratio prior ($r_{\text{prior}}$) and a width prior ($w_{\text{prior}}$). We model the joint probability distribution of nodule aspect ratio $r=h/w$ and log-width $\log(w)$ using a two-dimensional Gaussian Mixture Model with $M$ components.   

\begin{equation}
		p(r,\log(w))
		= \sum_{m=1}^{M} \pi_m\,
		\mathcal{N}\!\left(
		\begin{bmatrix}
			r\\[2pt]
			\log w
		\end{bmatrix};
		\boldsymbol{\mu}_m,\boldsymbol{\Sigma}_m
		\right)
		\label{eq:gmm_prior}
\end{equation}

where $\pi_m$ is the mixture weight, $\boldsymbol{\mu}_m$ and $\boldsymbol{\Sigma}_m$ denote the mean and covariance of the $m$-$th$ Gaussian component, $M$ is set to 3.
	
The resulting distribution of GMM is visualized in Fig. 3, 

where the contour map reveals a compact, elongated density region with a dominant mode around $r\approx1.0$ and moderate log-width values, followed by two smaller modes corresponding to relatively flat or slightly elongated nodules. This structured clustering supports our hypothesis that nodule geometry exhibits consistent statistical regularities across organs and devices. 
	
In the forward pass, a pair of prior parameters($r_{prior}$,$w_{prior}$) is sampled from the GMM and transformed into normalized factors $\tilde{r}_{\mathrm{prior}}$ and $\tilde{w}_{\mathrm{prior}}$. The network's predicted raw offset $\Delta P_{pred}=(\Delta x_{pred},\Delta y_{pred})$ is first scaled by the priors to conform to a statistically likely shape.

\begin{equation}
		\Delta x_{\mathrm{mod}}
		= \Delta x_{\mathrm{pred}}*\tilde{w}_{\mathrm{prior}},
		\qquad
		\Delta y_{\mathrm{mod}}
		= \Delta y_{\mathrm{pred}}*\tilde{w}_{\mathrm{prior}}*\tilde{r}_{\mathrm{prior}}
		\label{eq:prior_modulation}
\end{equation}
	
To enforce a hard constraint, the modulated offsets $\Delta p_{mod}=(\Delta x_{mod},\Delta y_{mod})$ are clamped within prior-defined boundaries. Let 
	
\begin{equation}
		max_x= \tilde{w}_{\mathrm{prior}},
		\qquad
		max_y= \tilde{w}_{\mathrm{prior}}*\tilde{r}_{\mathrm{prior}},
		\label{eq:prior_modulation}
\end{equation}
	
Then the final offsets are obtained as:
	
\begin{equation}
		\Delta x_{\text{final}} 
		= \text{clamp}\!\left( 
		\Delta x_{\text{mod}}, 
		-max_x, 
		max_x 
		\right)
		\label{eq:clamp_x}
\end{equation}
	
\begin{equation}
		\Delta y_{\text{final}} 
		= \text{clamp}\!\left( 
		\Delta y_{\text{mod}}, 
		-max_y, 
		max_y 
		\right)
		\label{eq:clamp_y}
\end{equation}

These steps constrain the sampling region to a prior-consistent box around each query position and transform deformable convolution from a purely data-driven operator into a prior-guided sampler driven by clinically observed morphology. The pseudocode detailing these operations is shown in Algorithm1.
    
\newcommand{\Step}[1]{\State \textbf{Step #1:}}
\begin{algorithm}[H]
\caption{Prior DCN}\label{alg:alg1}
	\begin{algorithmic}[1] 
		\State \textbf{Input:} Input feature map $X \in \mathbb{R}^{N \times L \times C}$, spatial dims $(H, W)$, GMM $\mathcal{G}(ratio, log(width))$.
		\State \textbf{Input:} Learnable params $W_o$.
		\State \textbf{Output:} Output feature map $Y \in \mathbb{R}^{N \times L \times C}$.
		\vspace{0.1cm}
		
		\Step{1: \textbf{Compute base offsets}}
		\State $\Delta P_{base} \leftarrow (\Delta P_{pred} + P_{grid}) \times \sigma(S_{pred}) \times S_{max}$
		
		\Step{2: \textbf{Apply GMM Prior Regularization}}
		\State $(r_{prior}, w_{prior}) \leftarrow \mathcal{G}.\text{sample}(G)$ 
		\State $\tilde{w}_{\mathrm{prior}} \leftarrow \text{Normalize}(\exp(w_{prior})).\text{expandAs}(\Delta P_{pred})$ // Get $\tilde{w}_{\mathrm{prior}}$
		\State $\tilde{r}_{\mathrm{prior}} \leftarrow r_{prior}.\text{expandAs}(\Delta P_{pred})$ // Get $\tilde{r}_{\mathrm{prior}}$
		
		\State $max_x \leftarrow \tilde{w}_{\mathrm{prior}}$ // Define x-boundary (Eq. 3)
		\State $\Delta x_{final} \leftarrow \text{clamp}(\Delta x_{mod}, -max_x, max_x)$ // Modulate \& Clamp x (Eq. 4)
		\State $max_y \leftarrow \tilde{w}_{\mathrm{prior}} \times \tilde{r}_{\mathrm{prior}}$ // Define y-boundary (Eq. 3)
		\State $\Delta y_{final} \leftarrow \text{clamp}(\Delta y_{mod}, -max_y, max_y)$ // Modulate \& Clamp y (Eq. 5)
		\State $\Delta P_{final} \leftarrow \text{stack}(\Delta x_{final}, \Delta y_{final})$
		
		\Step{3: \textbf{Apply Deformable Sampling and Output}}
		\State $X_{out} \leftarrow \text{DeformableSample}(X, \Delta P_{final})$ 
		\State $Y \leftarrow \text{Linear}(\text{Reshape}(X_{out}, (N, L, C)), W_o)$
		\State \Return $Y$
    \end{algorithmic}
\end{algorithm}	

Building on the Prior DCN, we design the Spatially-adaptive Deformable FFN with Prior Regularization (SDFPR) module. As illustrated in Fig. 3, SDFPR is composed of two core sub-modules: the Prior DCN Block and the Mix FFN. The Prior DCN block injects aspect ratio and width priors into the local receptive field and stabilizes offset learning for irregular nodules, while the Mix FFN models global semantic dependencies under adaptive normalization and DropPath regularization. This module is strategically embedded after the standard 3×3 convolution within each residual block of the ResNet50 backbone to progressively enhance prior-guided feature extraction for irregularly shaped nodules at multiple hierarchical stages.

\subsection{Structural Prior Modeling with Multi-scale Spatial-Frequency Feature Mixer}
Beyond geometric variability, ultrasound images are fundamentally shaped by frequency-dependent physical phenomena. Speckle noise appears predominantly as high-frequency fluctuations that obscure local boundaries, while acoustic shadowing and attenuation alter low-frequency components that encode global morphology. Motivated by these ultrasound-specific imaging characteristics, we design a Multi-scale Spatial-Frequency Feature Mixer (MSFFM) to extract contour and morphology priors by jointly processing features in the spatial and frequency domains across multiple scales. Specifically, MSFFM takes three backbone feature maps and refines them through a feature pyramid structure, where each level is processed by a Dual-Branch Feature Fusion Module (DBFFM). The DBFFM synergistically combines a Perception-Aggregation Spatial Convolution Branch for contour prior and a Frequency-Domain Fusion Branch for morphology prior, with their outputs adaptively weighted.

\begin{figure}[htbp]
    \centering
    \includegraphics[width=0.8\linewidth]{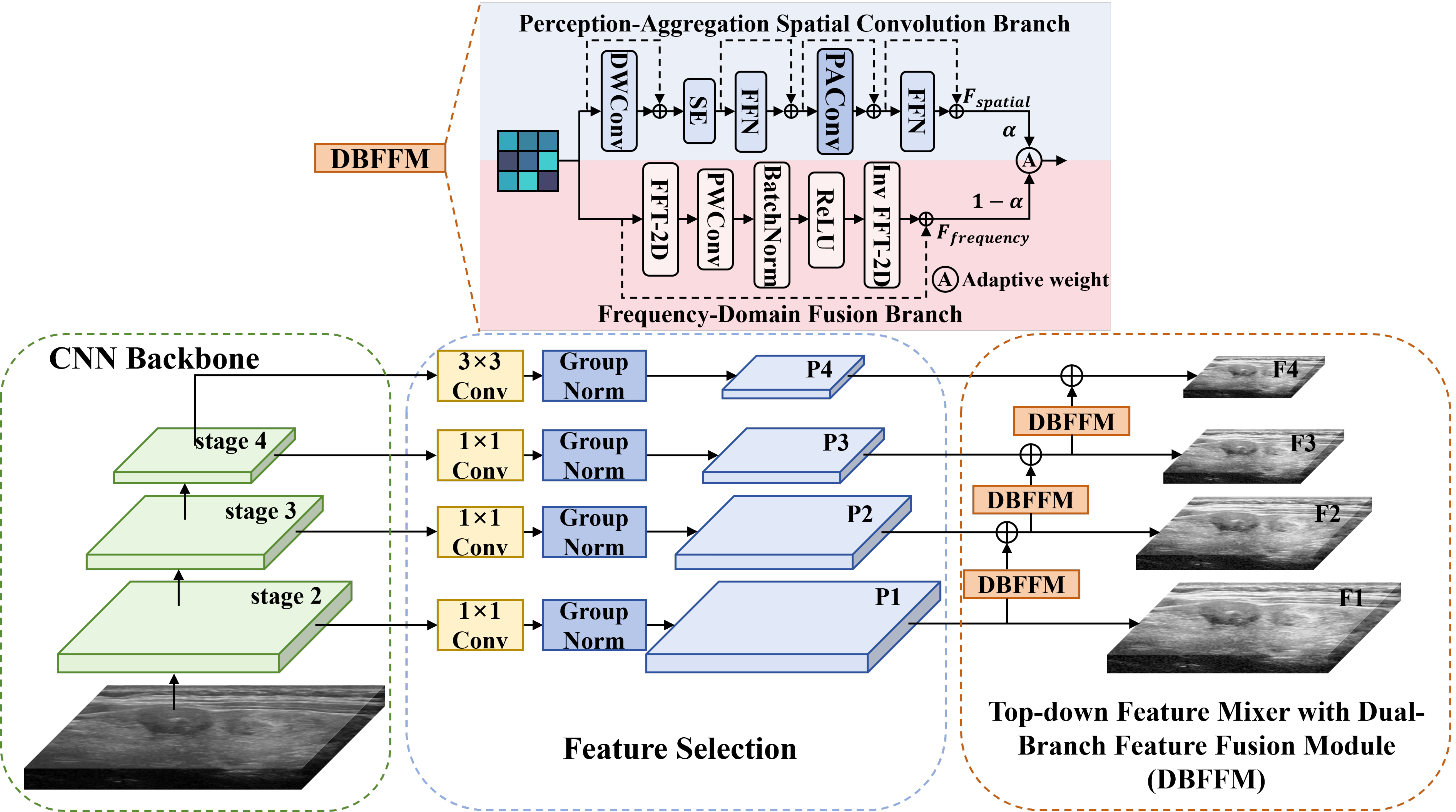} 
    \caption{Structure of the Multi-scale Spatial-Frequency Feature Mixer (MSFFM). MSFFM extracts structural priors by bridging two complementary domains: the Spatial Branch (top) aggregates local contour continuity via Perception-Aggregation Convolution, while the Frequency Branch (bottom) utilizes learnable spectral filtering to suppress speckle noise and highlight global morphology. An adaptive fusion strategy dynamically balances these local and global cues to generate robust representations across different nodule scales.} 
    \label{fig:fig4} 
\end{figure}

In the spatial domain, the Perception-Aggregation Convolution (PAConv) within the spatial branch enlarges the receptive field and aggregates context-aware information around nodules. As illustrated in Fig. 5(a), PAConv operates via a dual-strategy approach involving distinct perception and aggregation phases. The perception phase is responsible for acquiring comprehensive contextual information through a large receptive field and explicitly models spatial relationships across nodular regions. Subsequently, the aggregation phase adaptively fuses local features within highly correlated neighborhoods, thereby enhancing fine-grained visual representations and improving structural discriminability. These processes enable the network to learn structural priors such as boundary continuity, margin subtlety, and local homogeneity even under pronounced speckle interference. This process can be formally expressed as Eq. (6).           
	
\begin{equation}
		F_{PAConv}\ (x)=\mathcal{A}(\mathcal{P}(x_i,\mathcal{N}_P(x_i)),\mathcal{N}_A(x_i))
		\label{eq:F_fusion}
\end{equation}
	
where $\mathcal{N}_P(x_i))$ and $\mathcal{N}_A(x_i)$ denote the contextual regions for the perception and aggregation of the input token $x_i$, respectively, with $\mathcal{N}_P(x_i))$ encompassing a broader spatial extent than $\mathcal{N}_A(x_i)$.
	
Specifically, during the perception phase, a point-wise convolution (PWConv) is first employed to project visual tokens into a lower-dimensional embedding to reduce computational complexity. For each input feature $x_i$, a large-kernel depth-wise separable convolution (DWConv) with a kernel size of $K_p\times\ K_p$ is then applied to effectively capture the spatial contextual information within its perceptual neighborhood $\mathcal{N}_P(x_i))$. Subsequently, two additional PWConvs are applied to model the inter-token spatial dependencies, generating a context-adaptive weighting matrix $W \in \mathbb{R}^{H \times W \times D}$ that serves as the attention guidance for the subsequent aggregation step. During the aggregation phase, a grouped dynamic convolution is designed to efficiently integrate spatially correlated features. For the feature map $x_i \in \mathbb{R}^{H \times W \times C}$,the channel dimension is divided into $G$ groups, where all channels within the same group share convolutional weights to reduce memory consumption and computational cost. We reshape $w_i \in \mathbb{R}^D$ generated by the perception-phase into $w_i^{*} \in \mathbb{R}^{G \times K_a \times K_a}$. We then utilize $w_i^{*}$ to aggregate the highly correlated contextual region $\mathcal{N}_A(x_i)$.

\begin{figure}[htbp]
    \centering
    \includegraphics[width=0.8\linewidth]{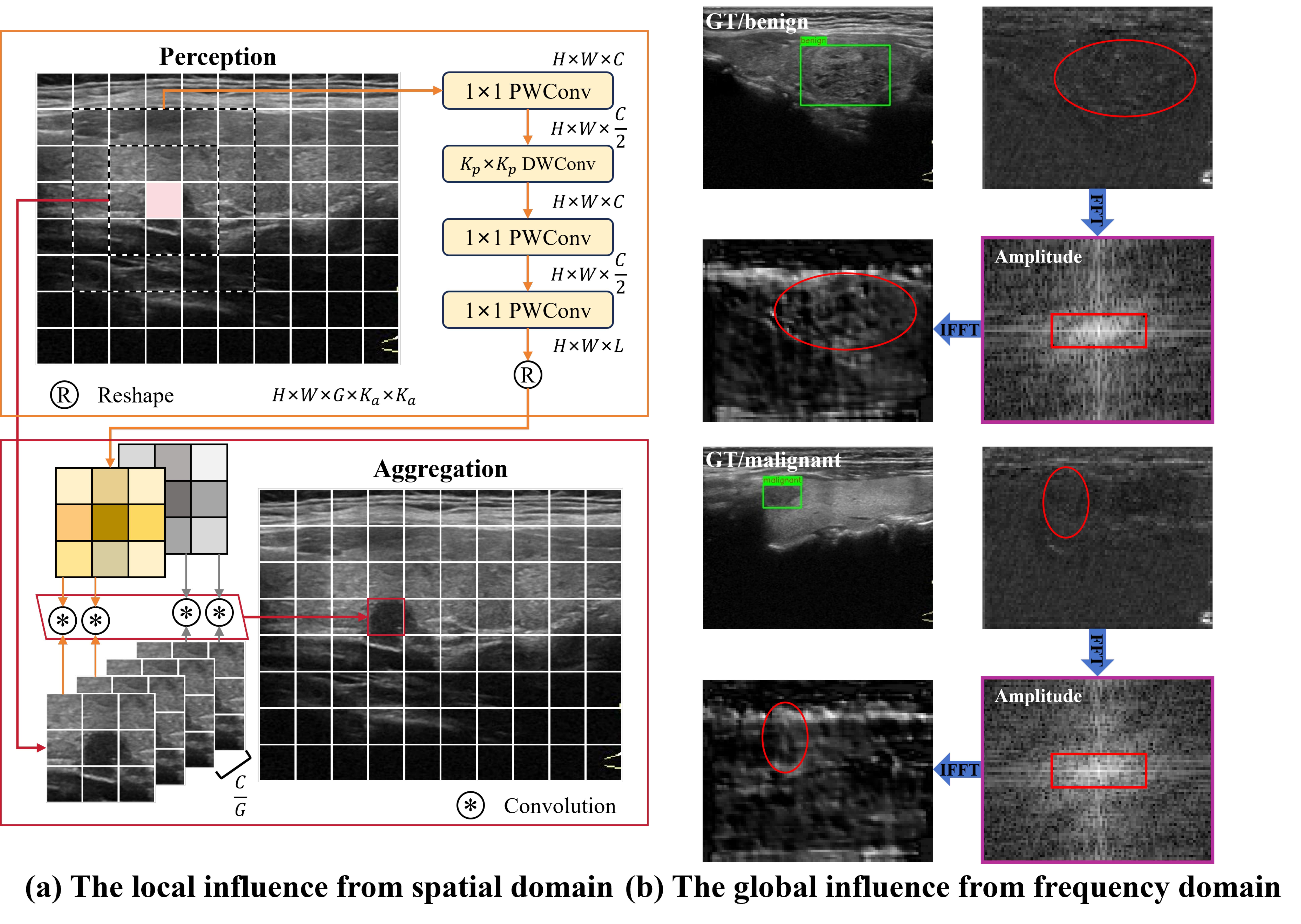} 
    \caption{(a) PAConv employs a dual-phase strategy to capture long-range spatial dependencies. (b) Frequency-Domain impact. Spatial features are transformed into the spectral domain, where learnable reweighting enhances morphology-related low-frequency components and suppresses speckle-dominated high-frequency noise. After inverse transformation, the reconstructed features exhibit cleaner backgrounds and more coherent nodule responses, reflecting a morphology prior.} 
    \label{fig:fig5} 
\end{figure}

This perception-aggregation spatial convolution branch further integrates DWConv, squeeze-and-excitation (SE) attention, and a FFN to improve feature representation. Skip connections are applied to stabilize training and preserve spatial details. 
	
In the frequency domain, the same multi-scale features are transformed by a two-dimensional FFT into their spectral representations. Low-frequency components in this domain describe coarse nodule shape and global tissue structure, whereas high-frequency bands are dominated by speckle and fine texture. The frequency branch applies PWConv, batch normalization, and ReLU on the amplitude spectrum as learnable filters that reweight individual frequency components according to their diagnostic importance. After this reweighting, an inverse FFT reconstructs a refined spatial feature map. As shown in Fig. 5(b), visual inspection shows that the reconstructed maps present cleaner backgrounds, stronger and more coherent responses around nodules, and reduced speckle-induced fluctuations. This behavior indicates that the frequency branch implicitly learns a morphology-aware spectral prior that separates tissue-related structure from physics-induced artifacts, which benefits ultrasound nodule detection especially when local intensity cues are unreliable.
	
The outputs of the spatial and frequency branches are then combined using an adaptive fusion strategy. For each scale, the final feature map is obtained as:
	
\begin{equation}
		\mathcal{F}(x) = (\alpha * F_{\text{spatial}}(x)) 
		\oplus ((1 - \alpha) * F_{\text{frequency}}(x))
		\label{eq:F_fusion}
\end{equation}
	
where $\alpha$ is a learnable scalar parameter initialized to 0.5 and optimized during training. This formulation allows the network to dynamically balance contour priors from the spatial domain and morphology priors from the frequency domain according to the current data characteristics, such as nodule size, shape complexity, and noise level. The adaptively fused multi-scale representations are flattened into an image feature sequence and passed to the Transformer encoder, which means that all subsequent stages operate on features that have already been jointly shaped by geometric priors from SDFPR and spatial-frequency structural priors from MSFFM.
	
\subsection{Prior-modulated Dense Feature Interaction Mechanism}

DETR-like models typically follow a standardized pipeline similar to the original DETR \cite{19carion2020end}. However, a critical limitation persists: these models exclusively leverage features from the final encoder layer as the Key and Value for decoder cross-attention, discarding rich multi-level semantic information from earlier layers. In the proposed framework, this limitation is more pronounced because encoder inputs have already been processed by SDFPR and MSFFM, so each encoder layer carries a distinct mixture of geometric priors and spatial-frequency structural priors. If only the last encoder layer is used, many prior information embedded in earlier layers may be underutilized. Beyond the "Original" approach (Fig. 6(a)), we also explored other direct one-to-one interaction strategies, including a "Sequential Mapping" (Fig. 6(b)) and a "Reversed Mapping" (Fig. 6(c)). However, as demonstrated in Sec. 4.4.5, these non-fused mappings are suboptimal and fail to properly integrate semantic context with spatial detail.

\begin{figure}[htbp]
    \centering
    \includegraphics[width=0.8\linewidth]{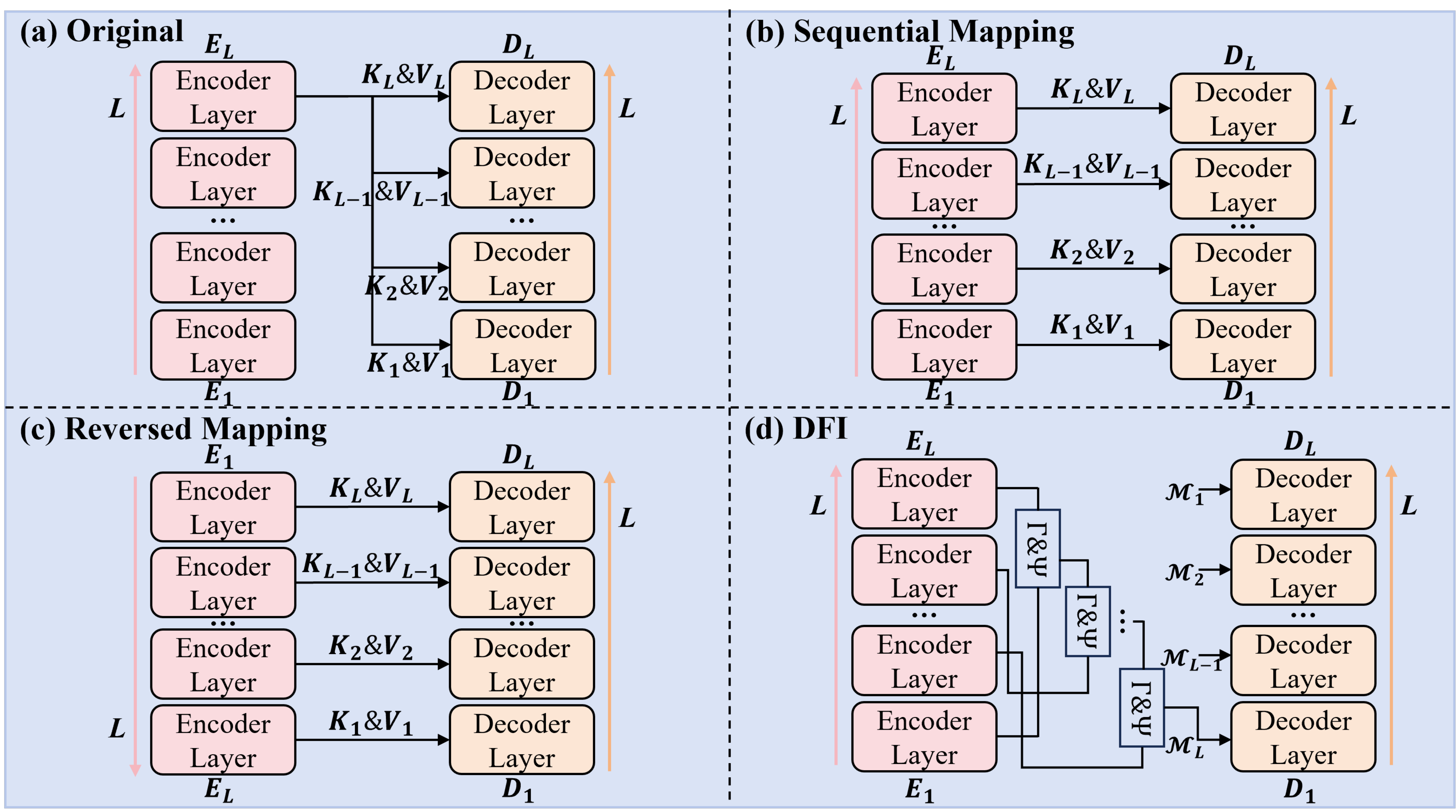} 
    \caption{Structure of the Dense Feature Interaction (DFI) Mechanism. DFI aggregates prior-enhanced features across all encoder layers and supplies them to decoder layers in a structured manner, ensuring that query refinement is consistently modulated by geometric and structural priors learned at different layers.} 
    \label{fig:fig6} 
\end{figure}
	
To fully exploit geometry- and structure-aware representations learned throughout the encoder, we propose a prior-modulated DFI mechanism (shown in Fig. 6(d)), inspired by the dense connections of DenseNet \cite{29huang2017densely}, to fully exploit features from different encoder layers in a way that preserves and propagates these multi-level priors into the decoding stage. It aggregates feature maps from all Transformer encoder layers ($E_1$ to $E_L$) in a top-down, iterative manner to produce a set of enhanced multi-level features ($\mathcal{M}_L$). These features are then supplied as the Key and Value to the decoder layers in a reverse order, ensuring that high-level semantic features guide early query refinement while low-level details inform later stages. The DFI mechanism establishes iterative cross-layer feature interaction across all encoder layers, ensuring that each layer's representation integrates both its intrinsic properties and complementary information from other layers. This design enables the decoder to access a more comprehensive feature space, thereby enhancing query refinement and detection robustness for morphologically complex nodules. The operation of multi-stage iterative interaction can be formulated as follows:
	
\begin{equation}
		\mathcal{M}_i \;=\; \Psi\!\Big( E^{(i)} \,\Vert\, 
		\bigoplus_{r>i} \Gamma_{r\to i}\big( E^{(r)} \big) \Big)
		\label{eq:Mi}
\end{equation}
	
where $E^{(i)}\in\mathbb{R}^{B\times L\times D}$ denotes the output of the $i$-th encoder layer, $i=1,2,\ldots,L$, where $L$ is the total number of encoder layers. 
$\Gamma_{r\to i}$ is a projection operator that aligns the feature spaces between layers $r$ and $i$. $\Vert$ denotes concatenation, $\bigoplus$ denotes feature aggregation, and $\Psi:\mathbb{R}^{2c}\to\mathbb{R}^{c}$ is a compression mapping.
	
These aggregated features are then used to define the Keys and Values for the decoder layers.
	
\begin{equation}
		K_j=V_j=\mathcal{M}_{\pi(j)}
		\label{eq:Mi}
\end{equation}
	
\begin{equation}
		\pi(j)=L-j+1,\ \ \ \ \ \ \ \ \ \ \ \ j=1,2,\ldots,L
		\label{eq:Mi}
\end{equation}
	
where $K_j$ and $V_j$ denote the Key and Value for the $j$-$th$ decoder layer, and the mapping $\pi(j)$ enforces a one-to-one correspondence between decoder and encoder layers in a reverse order. For example, when $L=6$, we have $K_1=V_1=\mathcal{M}_6$, $K_2=V_2=\mathcal{M}_5$, …, $K_6=V_6=\mathcal{M}_1$. Early decoder layers attend to features with strong global semantics and morphology priors, while later decoder layers work with features that retain more spatial detail and contour priors. This pairing ensures that both abstract prior context and fine structural information can be exploited at appropriate stages of query refinement.
	
The operation of the $j$-$th$ decoder layer can be formulated as follows:
	
\begin{equation}
		Q_{j} \;=\; \mathrm{FFN}_{j}\!\big( \mathrm{CrossAttn}_{j}\big(\mathrm{SelfAttn}_{j}(Q_{j-1}),\; K_{j},\; V_{j}\big)\big)
		\label{eq:Qj}
\end{equation} 
	
Here, $Q_{j-1}$ and $Q_j$ denote the input and output queries of the $j$-$th$ decoder layer, $\mathrm{SelfAttn}_{j}(\cdot)$ models intra-query dependencies, $\mathrm{CrossAttn}_{j}(\cdot)$ aligns the queries with DFI-enhanced image features, and $\mathrm{FFN}_{j}(\cdot)$ applies non-linear transformation and refinement.

\section{Experiments}
\subsection{Dataset Overview}
We evaluated the proposed framework on a total of 13,308 ultrasound images collected from four ultrasound nodule datasets, including two clinically collected thyroid detection datasets for internal validation, one public thyroid dataset for external validation, and one public breast dataset for generalization evaluation. Representative samples are shown in Fig. 7, and dataset statistics are summarized in Table I. 

\begin{figure}[htbp]
    \centering
    \includegraphics[width=0.8\linewidth]{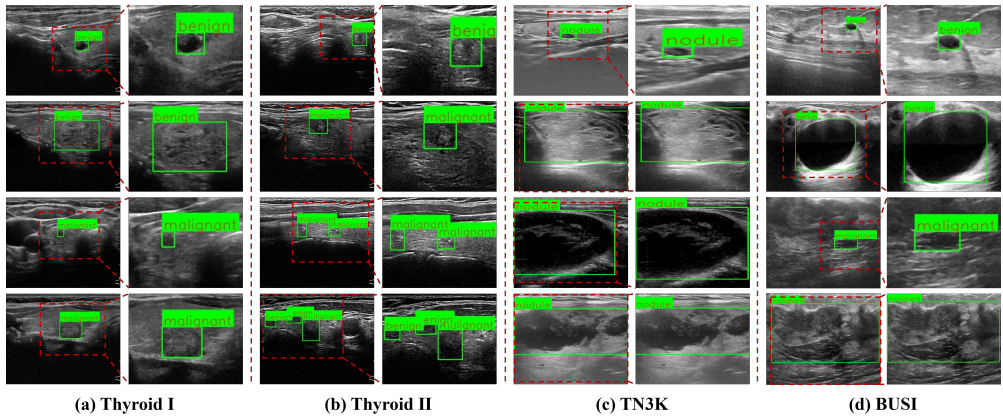} 
    \caption{Representative examples of ultrasound nodule images.} 
    \label{fig:fig7} 
\end{figure}

\begin{table*}[htbp]
\centering
\caption{Data distribution for each dataset, where BN denotes benign nodules and MN denotes malignant nodules.}
\begin{tabular}{lcccccc}
\toprule
			Dataset     & {Train} & {Val} & {Test} & {BN} & {MN} & {Total} \\
			\midrule
			Thyroid I   & 1151  & 352  & 346  & 327  & 1522 & 1849 \\
			Thyroid II  & 4089  & 1753 & 1459 & 1159 & 6646 & 7301 \\
			TN3K\cite{49gong2021multi}        & 2265  & 614  & 614  & {--} & {--} & 3493 \\
			BUSI\cite{50al2020dataset}        & 397   & 136  & 132  & 454  & 211  & 665  \\
\bottomrule
\end{tabular}
\vspace{1ex}
\begin{flushleft}
\end{flushleft}
\label{tab:CTV_comparison}
\end{table*}
	
\subsubsection{Private clinical datasets for internal validation}
Thyroid I and Thyroid II were collected at the Cancer Hospital of the Chinese Academy of Medical Sciences, using GE Logiq E9 and S7 systems between February 2018 and February 2019, under IRB approval (No. 24/243-4523). Together they include 9,150 thyroid ultrasound images (1,849 in Thyroid I and 7,301 in Thyroid II). All images were de-identified and annotated by senior radiologists, with benign/malignant labels derived from histopathology or structured reports. Compared with public datasets, these collections provide substantially larger sample sizes and broader case complexity, thereby better representing real-world clinical heterogeneity.
	
\subsubsection{Public thyroid dataset for external validation}
TN3K \cite{49gong2021multi} contains 3,493 Thyroid Images without benign–malignant labels and was originally designed for segmentation. Because their annotations are pixel-wise masks, we converted them to bounding boxes using a connected-component–based strategy.
	
\subsubsection{Public breast dataset for cross-organ generalization evaluation}
BUSI \cite{50al2020dataset} comprises 665 breast images, covering benign, malignant, and normal categories. Similar to TN3K, BUSI provides pixel-level segmentation masks, which were converted into bounding boxes using the same connected-component-based approach. Unlike thyroid datasets, breast nodules exhibit different anatomical contexts, background tissue characteristics, and shape distributions. Therefore, evaluation on BUSI enables a stringent assessment of whether the proposed prior-guided framework can generalize across organs.

\subsection{Experimental Setup}
\subsubsection{Implementation Details}
All experiments were conducted on an Ubuntu 18.04.6 LTS operating system with an Intel Xeon Gold 5118 CPU, 128 GB RAM, and an Nvidia RTX 3090 GPU (24GB). The framework was implemented in Python 3.8. For the model's architecture, we utilized ResNet50 as the backbone for feature extraction. The Transformer was configured with six encoder and six decoder layers, which we selected based on ablation findings demonstrating that this depth provided the best balance between accuracy and GPU memory constraints. A total of 300 object queries were used for the detection head. The model was trained for 200 epochs with a batch size of 2. The learning rate was set to 1e-4, combined with a weight decay of 1e-4, and positional encoding utilized a temperature coefficient of 20. For the loss function, we employed Focal Loss for classification, and a combination of L1 loss and Generalized Intersection over Union (GIoU) loss for bounding box regression.

\subsubsection{Evaluation Metrics}
We evaluated the model's performance using Average Precision ($AP$) under three thresholds: $AP$, $AP_{50}$ and $AP_{75}$. To analyze performance based on nodule scale, we also report metrics for small ($AP_s$), medium ($AP_m$), and large ($AP_l$) nodules, respectively. To further assess detection accuracy for different nodule types, $AP$@0.5-$BN$ and $AP$@0.5-$MN$ were employed to evaluate benign and malignant nodules, respectively. Here, $AP$ represents the area under the Precision-Recall ($PR$) curve.

\begin{equation}
		\mathrm{Precision}=\tfrac{\mathrm{TP}}{\mathrm{TP}+\mathrm{FP}}
		\label{eq:Precision}
\end{equation}
	
\begin{equation}
		\mathrm{Recall}=\tfrac{\mathrm{TP}}{\mathrm{TP}+\mathrm{FN}}
		\label{eq:Recall}
\end{equation}
	
\begin{equation}
		\mathrm{AP} \;=\; \int_{0}^{1} \mathrm{Precision}(R)\,\mathrm{d}R
		\label{eq:AP}
\end{equation}
	
In the above formulas, TP, FP, and FN denote true positives, false positives, and false negatives. A higher $AP$ value signifies superior model performance on the detection task.

\subsection{Comparative Experimental Results}
To comprehensively evaluate the effectiveness of the proposed method, we conducted comparison experiments against many representative detection frameworks, covering (i) CNN-based detectors, including Faster R-CNN \cite{12ren2016faster}, RetinaNet \cite{13lin2017focal}, YOLOv7 \cite{51wang2023yolov7}, YOLOv11 \cite{52khanam2024yolov11}, and YOLOv12 \cite{53tian2025yolov12}; (ii) DETR-based detectors, including DETR \cite{19carion2020end}, Deformable-DETR \cite{21zhu2020deformable}, DN-Deformable-DETR \cite{22li2022dn}, Lite-DETR \cite{54li2023lite}, Salience-DETR \cite{37hou2024salience}, and Deim \cite{38huang2025deim}; (iii) ultrasound-specific nodule detection methods, proposed by Liu et al. \cite{15liu2019automated}, Meng et al. \cite{65Meng2023}, Wang et al. \cite{56wang2023thinking} and Zhou et al. \cite{31zhou2024thyroid}; and (iv) a DETR-based model for leukocyte detection proposed by Chen et al. \cite{55chen2024accurate}, a RCNN-based model for caries detection proposed by Chen et al. \cite{66Chen2024} and a YOLO-based model for general medical detection proposed by Yu et al. \cite{67Yu2025}.
	
\subsubsection{Internal validation on private clinical datasets}
We first evaluated all methods on the two private clinical datasets, Thyroid I and Thyroid II. As shown in Table II and Table III, the proposed method outperforms all comparison approaches on both datasets. On Thyroid I, our method achieves the highest $AP$ and demonstrates clear advantages in detecting small and irregular nodules, which are particularly challenging due to blurred boundaries and speckle noise. Compared with Faster R-CNN, our method achieved improvements of 0.52 in $AP_s$, 0.259 in $AP_m$ and 0.12 in $AP_l$. Relative to the original DETR, it yielded gains of 0.328 in $AP_s$, 0.175 in $AP_m$ and 0.034 in $AP_l$, respectively. Furthermore, the class-wise $AP$ values reported in Table II enable more fine-grained analysis of model enhancements. Notably, our approach also outperformed the domain-specific methods of Liu et al. \cite{15liu2019automated}, Meng et al. \cite{65Meng2023}, Wang et al. \cite{56wang2023thinking}, Zhou et al. \cite{31zhou2024thyroid}, Chen et al. \cite{55chen2024accurate}, Chen et al. \cite{66Chen2024} and  Yu et al. \cite{67Yu2025}, underscoring the effectiveness of our multi-scale representation in thyroid nodule detection. On Thyroid I, the model attained the highest performance in terms of $AP$@0.5-BN (0.949), $AP$@0.5-MN (0.951), $AP$@0.5 (0.950), $AP$@0.75 (0.693) and $AP_l$ (0.709). Although its $AP$ (0.602), $AP_s$ (0.422) and $AP_m$ (0.608) were marginally lower than Lite-DETR, it achieved the best performance on all other $AP$ metrics.

\begin{table*}[htbp]
\centering
\caption{Comparison of different object detection models on Thyroid I dataset. '\textuparrow' indicates that higher values represent better performance. The current top three results are highlighted in \best{red}, \second{blue} and \third{green}.}
\begin{tabular}{lcccccccc}
\toprule
			Method 
			& {AP@0.5-BN$\uparrow$} & {AP@0.5-MN$\uparrow$} & {AP$\uparrow$} & {AP@0.5$\uparrow$} & {AP@0.75$\uparrow$} & {AP$_{s}\uparrow$} & {AP$_{m}\uparrow$} & {AP$_{l}\uparrow$} \\
			\midrule
			Faster-RCNN \cite{12ren2016faster}   & 0.928 & 0.802 & 0.438 & 0.865 & 0.392 & 0.107 & 0.408 & 0.663 \\
			RetinaNet \cite{13lin2017focal}        & 0.905 & \second{0.965} & \third{0.643} & 0.935 & 0.742 & 0.429 & 0.623 & 0.769 \\
			YOLOv7 \cite{51wang2023yolov7}          & 0.733 & 0.885 & 0.523 & 0.809 & 0.608 & 0.493 & 0.521 & 0.609 \\
			YOLOv11 \cite{52khanam2024yolov11}        & 0.841 & \third{0.964} & 0.608 & 0.902 & 0.739 & \third{0.555} & 0.573 & 0.620 \\
			YOLOv12 \cite{53tian2025yolov12}        & 0.898 & \best{0.967} & 0.596 & 0.933 & 0.738 & \third{0.555} & \third{0.628} & \third{0.789} \\
			\midrule
			DETR \cite{19carion2020end}            & 0.840 & 0.942 & 0.516 & 0.891 & 0.583 & 0.299 & 0.492 & 0.749 \\
			Deformable-DETR \cite{21zhu2020deformable}  & 0.878 & 0.939 & 0.509 & 0.908 & 0.516 & 0.266 & 0.525 & 0.535 \\
			DN-Deformable-DETR \cite{22li2022dn} & 0.922 & 0.963 & 0.621 & 0.943 & 0.706 & 0.452 & 0.601 & \best{0.793} \\
			Lite-DETR \cite{54li2023lite}   & 0.911 & 0.957 & 0.625 & 0.934 & 0.736 & 0.316 & 0.607 & \second{0.791} \\
			Salience-DETR \cite{37hou2024salience}   & 0.883 & \third{0.964} & 0.628 & 0.924 & 0.735 & 0.403 & \third{0.628} & 0.701 \\
			Deim \cite{38huang2025deim}   & 0.847 & \second{0.965} & 0.637 & 0.906 & 0.739 & 0.549 & 0.616 & 0.771 \\
			\midrule
			Liu et al. \cite{15liu2019automated}     & 0.810 & 0.931 & 0.565 & 0.870 & 0.679 & 0.549 & 0.552 & 0.652 \\
			Meng et al. \cite{65Meng2023}     & 0.363 & 0.838 & 0.369 & 0.600 & 0.415 & 0.059 & 0.379 & 0.710 \\
			Wang et al. \cite{56wang2023thinking} & \third{0.975} & 0.962 & 0.634 & \third{0.968} & \third{0.749} & \second{0.564} & 0.623 & 0.739 \\
			Zhou et al. \cite{31zhou2024thyroid}   & 0.912 & 0.946 & 0.621 & 0.929 & 0.729 & 0.475 & 0.603 & 0.785 \\
			\midrule
			Chen et al. \cite{55chen2024accurate}   & 0.957 & \third{0.964} & 0.625 & 0.960 & 0.715 & 0.549 & 0.601 & 0.777 \\
			Chen et al. \cite{66Chen2024}   & 0.905 & 0.913 & 0.575 & 0.909 & 0.676 & 0.545 & 0.572 & 0.648 \\
			Yu et al. \cite{67Yu2025}   & \second{0.987} & 0.959 & \second{0.653} & \second{0.973} & \second{0.768} & 0.430 & \second{0.646} & 0.755 \\
			\midrule
			Proposed    & \best{0.991} & \second{0.965} & \best{0.676} & \best{0.978} & \best{0.812} & \best{0.627} & \best{0.667} & 0.783 \\
\bottomrule
\end{tabular}
\vspace{1ex}
\begin{flushleft}
\end{flushleft}
\label{tab:CTV_comparison}
\end{table*}

\begin{table*}[htbp]
\centering
\caption{Comparison of different object detection models on Thyroid II dataset. '\textuparrow' indicates that higher values represent better performance. The current top three results are highlighted in \best{red}, \second{blue} and \third{green}.}
\begin{tabular}{lcccccccc}
\toprule
			Method 
			& {AP@0.5-BN$\uparrow$} & {AP@0.5-MN$\uparrow$} & {AP$\uparrow$} & {AP@0.5$\uparrow$} & {AP@0.75$\uparrow$} & {AP$_{s}\uparrow$} & {AP$_{m}\uparrow$} & {AP$_{l}\uparrow$} \\
			\midrule
			Faster-RCNN \cite{12ren2016faster}    & 0.907 & 0.819 & 0.430 & 0.863 & 0.354 & 0.117 & 0.424 & 0.603 \\
			RetinaNet \cite{13lin2017focal}       & 0.865 & 0.899 & 0.576 & 0.882 & 0.662 & 0.408 & 0.585 & 0.648 \\
			YOLOv7 \cite{51wang2023yolov7}         & 0.903 & 0.930 & 0.570 & 0.916 & 0.634 & 0.351 & 0.575 & 0.647 \\
			YOLOv11 \cite{52khanam2024yolov11}         & 0.881 & \second{0.949} & 0.577 & 0.915 & 0.627 & 0.366 & 0.545 & 0.566 \\
			YOLOv12 \cite{53tian2025yolov12}       & 0.886 & \best{0.951} & \third{0.597} & 0.919 & 0.642 & 0.389 & 0.592 & 0.680 \\
			\midrule 
			DETR \cite{19carion2020end}           & 0.702 & 0.911 & 0.448 & 0.806 & 0.444 & 0.254 & 0.444 & 0.597 \\
			Deformable-DETR \cite{21zhu2020deformable}  & 0.903 & 0.935 & 0.511 & 0.919 & 0.508 & 0.337 & 0.524 & 0.574 \\
			DN-Deformable-DETR \cite{22li2022dn}  & 0.916 & 0.937 & 0.575 & 0.926 & 0.622 & \third{0.421} & 0.580 & 0.656 \\
			Lite-DETR \cite{54li2023lite}  & 0.928 & \third{0.945} & \best{0.606} & 0.936 & \second{0.688} & \best{0.439} & \best{0.611} & \second{0.684} \\
			Salience-DETR \cite{37hou2024salience}   & 0.880 & 0.929 & 0.567 & 0.904 & 0.628 & 0.368 & 0.574 & 0.649 \\
			Deim \cite{38huang2025deim}  & 0.898 & 0.926 & 0.576 & 0.912 & 0.641 & 0.372 & 0.581 & 0.658 \\
			\midrule
			Liu et al. \cite{15liu2019automated}   & 0.866 & 0.899 & 0.522 & 0.882 & 0.575 & 0.390 & 0.534 & 0.558 \\
			Meng et al. \cite{65Meng2023}     & 0.558 & 0.763 & 0.363 & 0.660 & 0.361 & 0.152 & 0.369 & 0.464 \\
			Wang et al. \cite{56wang2023thinking}  & 0.908 & 0.927 & 0.586 & 0.918 & 0.668 & 0.414 & 0.587 & \third{0.682} \\
			Zhou et al. \cite{31zhou2024thyroid}   & 0.936 & 0.940 & 0.579 & 0.938 & 0.634 & 0.413 & 0.583 & 0.678 \\
			\midrule
			Chen et al. \cite{55chen2024accurate}  & \second{0.946} & 0.943 & 0.580 & \second{0.944} & 0.646 & 0.398 & 0.586 & 0.656 \\
			Chen et al. \cite{66Chen2024}  & 0.903 & 0.919 & 0.558 & 0.911 & 0.636 & 0.412 & 0.564 & 0.625 \\
			Yu et al. \cite{67Yu2025}   & \third{0.938} & 0.944 & 0.596 & \third{0.941} & \third{0.669} & 0.418 & \third{0.601} & 0.660 \\
			\midrule
			Proposed  & \best{0.949} & \best{0.951} & \second{0.602} & \best{0.950} & \best{0.693} & \second{0.422} & \second{0.608} & \best{0.709} \\
\bottomrule
\end{tabular}
\vspace{1ex}
\begin{flushleft}
\end{flushleft}
\label{tab:CTV_comparison}
\end{table*}

\begin{table*}[htbp]
\centering
\caption{Comparison of different object detection models on TN3K dataset. '\textuparrow' indicates that higher values represent better performance. The current top three results are highlighted in \best{red}, \second{blue} and \third{green}.}
\begin{tabular}{lcccccc}
\toprule
			Method 
			& {AP$\uparrow$} & {AP@0.5$\uparrow$} & {AP@0.75$\uparrow$} & {AP$_{s}\uparrow$} & {AP$_{m}\uparrow$} & {AP$_{l}\uparrow$} \\
			\midrule
			Faster-RCNN \cite{12ren2016faster}   & 0.492 & \second{0.863} & 0.513 & 0.065 & 0.377 & 0.556 \\
			RetinaNet \cite{13lin2017focal}       & 0.474 & 0.790 & 0.506 & 0.191 & 0.424 & 0.525 \\
			YOLOv7 \cite{51wang2023yolov7}         & 0.500 & 0.821 & 0.550 & 0.241 & 0.450 & 0.574 \\
			YOLOv11 \cite{52khanam2024yolov11}        & \second{0.529} & 0.848 & 0.565 & 0.247 & 0.445 & 0.517 \\
			YOLOv12 \cite{53tian2025yolov12}       & \third{0.526} & 0.848 & 0.567 & 0.241 & 0.447 & 0.576 \\
			\midrule
			DETR \cite{19carion2020end}          & 0.462 & 0.823 & 0.467 & 0.137 & 0.404 & 0.532 \\
			Deformable-DETR \cite{21zhu2020deformable}  & 0.407 & 0.791 & 0.365 & 0.103 & 0.355 & 0.470 \\
			DN-Deformable-DETR \cite{22li2022dn} & 0.493 & 0.816 & 0.522 & 0.235 & 0.402 & 0.563 \\
			Lite-DETR \cite{54li2023lite}   & 0.508 & 0.827 & 0.554 & \second{0.278} & \second{0.465} & 0.559 \\
			Salience-DETR \cite{37hou2024salience}  & 0.524 & 0.826 & \third{0.572} & \third{0.276} & \third{0.452} & \second{0.586} \\
			Deim \cite{38huang2025deim}  & 0.467 & 0.715 & 0.531 & 0.167 & 0.420 & 0.522 \\
			\midrule
			Liu et al. \cite{15liu2019automated}     & 0.467 & 0.820 & 0.485 & 0.084 & 0.402 & 0.521 \\
			Meng et al. \cite{65Meng2023}     & 0.214 & 0.387 & 0.213 & 0.106 & 0.194 & 0.243 \\
			Wang et al. \cite{56wang2023thinking} & 0.456 & 0.802 & 0.454 & 0.194 & 0.415 & 0.510 \\
			Zhou et al. \cite{31zhou2024thyroid}  & 0.475 & 0.815 & 0.514 & 0.234 & 0.425 & 0.536 \\
			\midrule
			Chen et al. \cite{55chen2024accurate}   & 0.510 & \third{0.852} & 0.550 & 0.238 & 0.445 & \third{0.582} \\
			Chen et al. \cite{66Chen2024}  & 0.476 & 0.830 & 0.505 & 0.069 & 0.431 & 0.520 \\
			Yu et al. \cite{67Yu2025}   & 0.516 & 0.843 & \second{0.576} & 0.249 & \third{0.452} & 0.559 \\
			\midrule
			Proposed     & \best{0.540} & \best{0.864} & \best{0.605} & \best{0.280} & \best{0.469} & \best{0.604} \\
\bottomrule
\end{tabular}
\vspace{1ex}
\begin{flushleft}
\end{flushleft}
\label{tab:CTV_comparison}
\end{table*}

\begin{table*}[htbp]
\centering
\caption{Comparison of different object detection models on BUSI dataset. '\textuparrow' indicates that higher values represent better performance. The current top three results are highlighted in \best{red}, \second{blue} and \third{green}.}
\begin{tabular}{lcccccccc}
\toprule
			Method 
			& {AP@0.5-BN$\uparrow$} & {AP@0.5-MN$\uparrow$} & {AP$\uparrow$} & {AP@0.5$\uparrow$} & {AP@0.75$\uparrow$} & {AP$_{s}\uparrow$} & {AP$_{m}\uparrow$} & {AP$_{l}\uparrow$} \\
			\midrule
			Faster-RCNN \cite{12ren2016faster}   & 0.689 & \best{0.693} & 0.413 & \third{0.691} & 0.464 & 0.200 & 0.229 & \third{0.438} \\
			RetinaNet \cite{13lin2017focal}      & 0.714 & 0.592 & 0.398 & 0.653 & 0.401 & 0.200 & 0.329 & 0.396 \\
			YOLOv7 \cite{51wang2023yolov7}          & 0.711 & 0.429 & 0.356 & 0.570 & 0.399 & \best{0.600} & 0.265 & 0.337 \\
			YOLOv11 \cite{52khanam2024yolov11}        & 0.706 & 0.468 & 0.365 & 0.587 & 0.388 & \best{0.600} & 0.316 & 0.288 \\
			YOLOv12 \cite{53tian2025yolov12}        & \best{0.783} & 0.543 & \third{0.454} & 0.663 & 0.478 & \best{0.600} & \third{0.357} & 0.405 \\
			\midrule
			DETR \cite{19carion2020end}           & \second{0.749} & 0.630 & 0.423 & 0.689 & 0.461 & 0.300 & 0.274 & 0.433 \\
			Deformable-DETR \cite{21zhu2020deformable}  & 0.654 & 0.491 & 0.388 & 0.573 & 0.454 & \third{0.476} & 0.308 & 0.379 \\
			DN-Deformable-DETR \cite{22li2022dn} & 0.714 & 0.618 & 0.441 & 0.666 & 0.463 & 0.364 & 0.302 & 0.437 \\
			Lite-DETR \cite{54li2023lite}  & \third{0.745} & 0.570 & \second{0.455} & 0.657 & \third{0.487} & 0.044 & 0.340 & \second{0.447} \\
			Salience-DETR \cite{37hou2024salience}  & 0.709 & 0.561 & 0.416 & 0.635 & 0.417 & 0.080 & 0.308 & 0.421 \\
			Deim \cite{38huang2025deim}  & 0.672 & 0.500 & 0.435 & 0.586 & \second{0.502} & 0.350 & \second{0.373} & 0.416 \\
			\midrule
			Liu et al. \cite{15liu2019automated}  & 0.717 & 0.573 & 0.406 & 0.645 & 0.451 & \second{0.500} & 0.263 & 0.419 \\
			Meng et al. \cite{65Meng2023}    & 0.741 & \third{0.649} & 0.391 & \second{0.695} & 0.389 & 0.044 & 0.356 & 0.404 \\
			Wang et al. \cite{56wang2023thinking} & 0.681 & 0.587 & 0.406 & 0.634 & 0.456 & 0.400 & 0.299 & 0.393 \\
			Zhou et al. \cite{31zhou2024thyroid}  & 0.738 & 0.534 & 0.423 & 0.636 & 0.458 & 0.400 & 0.296 & 0.421 \\
			\midrule
			Chen et al. \cite{55chen2024accurate}  & 0.742 & 0.556 & 0.394 & 0.649 & 0.416 & \second{0.500} & 0.295 & 0.395 \\
			Chen et al. \cite{66Chen2024}  & 0.679 & 0.505 & 0.305 & 0.592 & 0.292 & \second{0.500} & 0.232 & 0.307 \\
			Yu et al. \cite{67Yu2025}  & 0.633 & 0.591 & 0.384 & 0.612 & 0.421 & \best{0.600} & 0.331 & 0.350 \\
			\midrule
			Proposed     & 0.744 & \second{0.668} & \best{0.472} & \best{0.706} & \best{0.585} & \best{0.600} & \best{0.389} & \best{0.470} \\
\bottomrule
\end{tabular}
\vspace{1ex}
\begin{flushleft}
\end{flushleft}
\label{tab:CTV_comparison}
\end{table*}

\subsubsection{External validation on the public thyroid dataset TN3K}
As reported in Table IV, our model achieved first-place performance across all metrics, including $AP$ (0.540), $AP$@0.5 (0.864), $AP$@0.75 (0.605), $AP_s$ (0.280), $AP_m$ (0.469) and $AP_l$ (0.604), proving its state-of-the-art capability. These results indicate that the geometric and structural priors learned by the proposed model are not overfitted to a specific clinical center but can generalize effectively to independent thyroid ultrasound data.
	
\subsubsection{Cross-organ generalization on the public breast dataset BUSI}
Finally, we evaluated cross-organ generalization performance on the BUSI dataset to investigate whether the proposed framework can transfer learned priors across different anatomical sites. As shown in Table V, our method again achieved top-ranked performance across all AP metrics, including $AP$ (0.472), $AP$@0.5 (0.706), $AP$@0.75 (0.585), $AP_s$ (0.600), $AP_m$ (0.389) and $AP_l$ (0.470). This comprehensive superiority is particularly noteworthy when interpreted in conjunction with the statistical size distribution in Fig. 1. As shown, large nodules form the majority in both TN3K and BUSI, and our model consistently achieved the best $AP_l$ scores on these datasets (0.604 on TN3K, 0.470 on BUSI) further confirm that the proposed method is particularly effective in accurately localizing and classifying large-scale ultrasound nodules. More importantly, although small nodules constitute only a very limited fraction in BUSI, our method still attained the highest $AP_s$ (0.600), demonstrating that the proposed multi-scale spatial–frequency representation and dense feature interaction design effectively preserve robustness even under severe data imbalance.   

\subsection{Ablation Experiments}\label{formats}
In this section, we conducted a comprehensive ablation study to verify the effectiveness of each proposed component. In addition, we also performed in-depth research on the internal structure of the module to determine the most appropriate hyperparameters. It is worth noting that all ablation experiments were conducted on the Thyroid I dataset.
	
\subsubsection{Effectiveness of Each Designed Component}
To evaluate the individual contributions of the proposed modules from a prior-integration perspective, we conducted ablation experiments on the Thyroid I dataset by selectively enabling or disabling geometric, structural, and interaction-level priors into the baseline detector. The quantitative results are summarized in Table VI.
	
The baseline model, which does not incorporate any explicit prior modeling, achieved an $AP$ of 0.612, $AP_s$ of 0.429, $AP_m$ of 0.605 and $AP_l$ of 0.749. Incorporating the SDFPR module alone significantly improved overall detection accuracy ($AP$ = 0.642), highlighting geometric prior's effectiveness in capturing irregular shapes and blurred boundaries. The MSFFM module also yielded notable gains ($AP$ = 0.635), particularly enhancing small-object detection ($AP_s$ = 0.569), which validates the importance of integrating spatial features with frequency-domain filtering that attenuates speckle-dominated noise and preserves structural cues across scales. Meanwhile, introducing the DFI improved detection consistency across scales ($AP$ = 0.627), demonstrating the benefit of multi-level query refinement.
	
When combining modules, further improvements were observed. The joint integration of SDFPR and MSFFM led to an $AP$ of 0.642, while SDFPR and DFI improved large-object detection ($AP_l$ = 0.796). The combination of MSFFM and DFI also enhanced performance across scales, achieving an $AP$ of 0.643. The full model, incorporating all three modules, achieved the best results, with an $AP$ of 0.676, $AP$@0.5 of 0.978, $AP$@0.75 of 0.812, and consistently strong performance across small, medium, and large nodules.
	
Overall, these results confirm how geometric priors, structural priors, and prior-modulated feature interaction jointly enhance detection accuracy under challenging ultrasound conditions. Their synergistic integration is essential for achieving the state-of-the-art performance of the proposed method.

\begin{table*}[htbp]
\centering
\caption{Ablation study of the proposed modules on Thyroid I dataset.}
\begin{tabular}{lcccccccccc}
\toprule
			SDFPR & MSFFM & DFI
			& {AP@0.5-BN$\uparrow$} & {AP@0.5-MN$\uparrow$} & {AP$\uparrow$} & {AP@0.5$\uparrow$} & {AP@0.75$\uparrow$} & {AP$_{s}\uparrow$} & {AP$_{m}\uparrow$} & {AP$_{l}\uparrow$} \\
			\midrule
			& & & 0.898 & 0.965 & 0.612 & 0.932 & 0.717 & 0.429 & 0.605 & 0.749 \\
			$\surd$      & & & 0.958 & 0.962 & 0.642 & 0.960 & 0.729 & 0.369 & 0.631 & 0.781 \\
			& $\surd$      & & 0.956 & 0.961 & 0.635 & 0.958 & 0.738 & 0.569 & 0.623 & 0.743 \\
			& & $\surd$      & 0.914 & 0.965 & 0.627 & 0.939 & 0.727 & 0.417 & 0.611 & 0.776 \\
			$\surd$      & $\surd$      & & $\underline{0.984}$ & 0.966 & 0.642 & 0.975 & 0.740 & 0.547 & $\underline{0.635}$ & 0.763 \\
			$\surd$      & & $\surd$      & 0.955 & $\underline{0.968}$ & 0.622 & 0.961 & 0.709 & 0.396 & 0.605 & $\mathbf{0.796}$ \\
			& $\surd$      & $\surd$      & 0.975 & $\mathbf{0.977}$ & $\underline{0.643}$ & \underline{0.976} & \underline{0.741} & \underline{0.576} & 0.628 & 0.751 \\
			$\surd$      & $\surd$      & $\surd$      & $\mathbf{0.991}$ & 0.965 & $\mathbf{0.676}$ & $\mathbf{0.978}$ & $\mathbf{0.812}$ & $\mathbf{0.627}$ & $\mathbf{0.667}$ & \underline{0.783} \\
\bottomrule
\end{tabular}
\vspace{1ex}
\begin{flushleft}
\end{flushleft}
\label{tab:CTV_comparison}
\end{table*}

\subsubsection{Effectiveness of Prior DCN in SDFPR}
To specifically validate the effectiveness of our proposed Prior DCN (Fig. 3), we conducted an ablation study on the Thyroid I dataset. We compared our full model, which uses the Prior DCN, against a variant where the Prior DCN within the SDFPR module was replaced with the original DCNv4.
	
The quantitative results are summarized in Table VII. The findings clearly indicate that the model incorporating our Prior DCN achieves superior performance across most evaluation metrics compared to the variant using the standard DCNv4. Notably, our approach demonstrates significant gains in $AP$ (+0.034), $AP$@0.75 (+0.054), $AP_s$ (+0.061) and $AP_m$ (+0.034). These results indicate that introducing aspect ratio and width priors effectively stabilizes offset learning and enhances the adaptability of deformable sampling to irregular and anisotropic nodule morphology.

\begin{table*}[htbp]
\centering
\caption{Ablation study comparing Prior DCN with the original DCNv4 on the Thyroid I dataset.}
\begin{tabular}{lcccccccccc}
\toprule
			Method & {AP@0.5-BN$\uparrow$} & {AP@0.5-MN$\uparrow$} & {AP$\uparrow$} & {AP@0.5$\uparrow$} & {AP@0.75$\uparrow$} & {AP$_{s}\uparrow$} & {AP$_{m}\uparrow$} & {AP$_{l}\uparrow$} \\
			\midrule
			w/ DCNv4 & 0.979 & {$\mathbf{0.973}$} & 0.642 & 0.976 & 0.758 & 0.566 & 0.633 & 0.773 \\
			w/ Prior DCN & {$\mathbf{0.991}$} & 0.965 & {$\mathbf{0.676}$} & {$\mathbf{0.978}$} & {$\mathbf{0.812}$} & {$\mathbf{0.627}$} & {$\mathbf{0.667}$} & {$\mathbf{0.783}$} \\
\bottomrule
\end{tabular}
\vspace{1ex}
\begin{flushleft}
\end{flushleft}
\label{tab:CTV_comparison}
\end{table*}
	
\subsubsection{Ablation on the DBFFM Fusion Strategy in MSFFM}
To validate the parallel, adaptively-weighted fusion strategy of the DBFFM within our MSFFM (Fig. 4), we conducted an ablation study on the Thyroid I dataset, with quantitative results presented in Table VIII. We compared our full parallel DBFFM against four alternatives: using only the Spatial branch, only the Freq branch, a serial Spatial $\rightarrow$ Freq cascade, and a serial Freq $\rightarrow$ Spatial cascade. The single-branch ablations, Spatial-only (0.644$AP$) and Freq-only (0.642$AP$), performed similarly, establishing a baseline and confirming that both spatial (local) and frequency (global) information are necessary, as both are significantly outperformed by any fusion strategy. Serializing the branches yielded improvements, with the Freq $\rightarrow$ Spatial configuration (0.667$AP$) notably outperforming the Spatial $\rightarrow$ Freq (0.650$AP$), suggesting that global filtering followed by local refinement is a viable approach. Our parallel DBFFM, however, achieved the highest overall performance (0.676$AP$), demonstrating its superiority. The most critical advantage is revealed in the high-IoU precision metric: our parallel DBFFM achieved an $AP$@0.75 of 0.812, representing a substantial leap over the next-best serial Freq $\rightarrow$ Spatial (0.786) and far exceeding the Spatial-only (0.710) and Freq-only (0.751) branches. This significant +0.026 to +0.102 gain in $AP$@0.75 strongly indicates that a serial pipeline creates an "information bottleneck" that degrades complementary features. Only the parallel, adaptively weighted architecture allows the network to simultaneously process fine-grained local textures and global morphological context.

\begin{table*}[htbp]
\centering
\caption{Ablation study on the DBFFM Fusion Strategy in MSFFM on the Thyroid I dataset.}
\begin{tabular}{lcccccccccc}
\toprule
			Method & {AP@0.5-BN$\uparrow$} & {AP@0.5-MN$\uparrow$} & {AP$\uparrow$} & {AP@0.5$\uparrow$} & {AP@0.75$\uparrow$} & {AP$_{s}\uparrow$} & {AP$_{m}\uparrow$} & {AP$_{l}\uparrow$} \\
			\midrule
			Spatial & {$\underline{0.984}$} & 0.952 & 0.644 & 0.968 & 0.710 & 0.555 & 0.626 & {$\mathbf{0.790}$} \\
			Freq & 0.965 & {$\underline{0.969}$} & 0.642 & 0.967 & 0.751 & 0.546 & 0.637 & 0.779 \\
			Spatial $\rightarrow$ Freq & 0.980 & 0.962 & 0.650 & 0.971 & 0.781 & 0.576 & 0.643 & 0.758 \\
			Freq $\rightarrow$ Spatial & 0.981 & {$\mathbf{0.976}$} & {$\underline{0.667}$} & {$\mathbf{0.979}$} & {$\underline{0.786}$} & {$\underline{0.617}$} & {$\underline{0.661}$} & 0.763 \\
			DBFFM & {$\mathbf{0.991}$} & 0.965 & {$\mathbf{0.676}$} & {$\underline{0.978}$} & {$\mathbf{0.812}$} & {$\mathbf{0.627}$} & {$\mathbf{0.667}$} & {$\underline{0.783}$} \\
\bottomrule
\end{tabular}
\vspace{1ex}
\begin{flushleft}
\end{flushleft}
\label{tab:CTV_comparison}
\end{table*}
	
\subsubsection{Analysis of the DFI Mechanism}
To validate our DFI mechanism (Fig. 6(d)), we conducted an ablation study on the Thyroid I dataset (shown in Table IX) comparing four encoder-decoder interaction strategies. The "Original" baseline (Fig. 6(a)), using only the $E_L$ layer, achieved 0.642 $AP$. A "Sequential Mapping" (Fig. 6(b), $E_1\rightarrow\ D_1$, …, $E_L\rightarrow\ D_L$) was detrimental to performance, causing a significant $AP$ drop to 0.591. This result confirms that initial decoder layers require strong semantic guidance, which the high-resolution but semantically ambiguous $E_1$ layer cannot provide. Furthermore, a simple "Reversed Mapping" (Fig. 6(c), $E_L\rightarrow\ D_1$, …, $E_1\rightarrow\ D_L$) also failed, degrading $AP$ to 0.605. This demonstrates that while final decoder layers ($D_L$) require high-resolution features for refinement, the raw $E_1$ features are ambiguous; the cross-attention mechanism struggles to differentiate salient nodule boundaries from high-frequency artifacts or tissue textures. In contrast, our DFI, which pairs this reversed mapping with dense feature fusion, achieved the top $AP$ of 0.676.

\begin{table*}[htbp]
\centering
\caption{Ablation study of the DFI Mechanism on the Thyroid I dataset.}
\begin{tabular}{lcccccccccc}
\toprule
			Method & {AP@0.5-BN$\uparrow$} & {AP@0.5-MN$\uparrow$} & {AP$\uparrow$} & {AP@0.5$\uparrow$} & {AP@0.75$\uparrow$} & {AP$_{s}\uparrow$} & {AP$_{m}\uparrow$} & {AP$_{l}\uparrow$} \\
			\midrule
			Original & {$\underline{0.984}$} & {$\underline{0.966}$} & {$\underline{0.642}$} & {$\underline{0.975}$} & {$\underline{0.740}$} & 0.547 & {$\underline{0.635}$} & 0.763 \\
			Sequential Mapping & 0.946 & 0.945 & 0.591 & 0.945 & 0.669 & 0.533 & 0.583 & 0.721 \\
			Reversed Mapping & 0.937 & {$\mathbf{0.970}$} & 0.605 & 0.953 & 0.674 & {$\underline{0.610}$} & 0.581 & {$\underline{0.765}$} \\
			DFI & {$\mathbf{0.991}$} & 0.965 & {$\mathbf{0.676}$} & {$\mathbf{0.978}$} & {$\mathbf{0.812}$} & {$\mathbf{0.627}$} & {$\mathbf{0.667}$} & {$\mathbf{0.783}$} \\
\bottomrule
\end{tabular}
\vspace{1ex}
\begin{flushleft}
\end{flushleft}
\label{tab:CTV_comparison}
\end{table*}
	
\subsubsection{Analysis of Hyperparameter Settings}
In addition to validating the proposed method, we systematically analyzed two critical hyperparameters: the depth of the Transformer and the composition of the loss function.
	
\textbf{Transformer Encoder and Decoder Depth Analysis.} In DETR-like architectures, the number of encoder and decoder layers strongly influences detection performance. Based on insights from existing studies \cite{57zhao2024detrs} \cite{58yao2021efficient}, we selected several promising ablation configurations, as summarized in Table X. The results demonstrate that deeper models generally achieve better metrics but incur substantially higher computational and memory costs. Considering the limitations of our hardware, we identified the configuration with six encoder layers and six decoder layers as the optimal choice.

\begin{table*}[htbp]
\centering
\caption{Ablation study with different numbers of encoder layers and decoder layers on Thyroid I dataset.}
\begin{tabular}{lcccccccccc}
\toprule
			Encoder & Decoder & {AP@0.5-BN$\uparrow$} & {AP@0.5-MN$\uparrow$} & {AP$\uparrow$} & {AP@0.5$\uparrow$} & {AP@0.75$\uparrow$} & {AP$_{s}\uparrow$} & {AP$_{m}\uparrow$} & {AP$_{l}\uparrow$} \\
			\midrule
			1 & 1 & 0.834 & 0.874 & 0.470 & 0.854 & 0.461 & 0.248 & 0.453 & 0.646 \\
			3 & 1 & 0.807 & 0.908 & 0.474 & 0.857 & 0.495 & 0.201 & 0.453 & 0.673 \\
			6 & 1 & 0.764 & 0.894 & 0.484 & 0.829 & 0.543 & 0.475 & 0.454 & 0.644 \\
			3 & 3 & 0.942 & {$\mathbf{0.971}$} & {$\underline{0.618}$} & {$\underline{0.957}$} & 0.667 & 0.496 & {$\underline{0.597}$} & {$\mathbf{0.788}$} \\
			6 & 3 & {$\underline{0.944}$} & {$\underline{0.969}$} & 0.611 & 0.956 & {$\underline{0.701}$} & {$\underline{0.574}$} & 0.592 & 0.755 \\
			6 & 6 & {$\mathbf{0.991}$} & 0.965 & {$\mathbf{0.676}$} & {$\mathbf{0.978}$} & {$\mathbf{0.812}$} & {$\mathbf{0.627}$} & {$\mathbf{0.667}$} & {$\underline{0.767}$} \\
\bottomrule
\end{tabular}
\vspace{1ex}
\begin{flushleft}
\end{flushleft}
\label{tab:CTV_comparison}
\end{table*}
	
\textbf{Loss Function Composition Analysis.} Our training objective combines Focal loss for classification with L1 and GIoU losses for bounding box regression. As presented in Table XI, when all three losses (Focal + L1 + GIoU) were used, the model achieved the best performance across most metrics, including $AP$@0.5-BN: 0.991, $AP$@0.5-MN: 0.965, $AP$: 0.676, $AP$@0.5: 0.978, $AP$@0.75: 0.812, $AP_s$: 0.627, and $AP_m$: 0.667. Removing L1 loss led to a slight performance drop across all metrics, highlighting its importance for precise localization. Excluding GIoU loss caused a more significant performance decline in $AP_s$, indicating that spatial accuracy suffers more noticeably in the absence of GIoU supervision. These findings support retaining both L1 and GIoU to achieve a balanced and effective optimization of classification and regression.

\begin{table*}[htbp]
\centering
\caption{Ablation study with different combinations of loss functions on Thyroid I dataset.}
\begin{tabular}{lcccccccccc}
\toprule
			Focal & L1 & GIoU & {AP@0.5-BN$\uparrow$} & {AP@0.5-MN$\uparrow$} & {AP$\uparrow$} & {AP@0.5$\uparrow$} & {AP@0.75$\uparrow$} & {AP$_{s}\uparrow$} & {AP$_{m}\uparrow$} & {AP$_{l}\uparrow$} \\
			\midrule
			$\surd$ & $\surd$ &  & {$\underline{0.972}$} & {$\underline{0.953}$} & 0.608 & {$\underline{0.962}$} & {$\underline{0.697}$} & {$\underline{0.553}$} & 0.589 & {$\mathbf{0.787}$} \\
			$\surd$ & & $\surd$ & 0.971 & 0.951 & {$\underline{0.616}$} & 0.961 & 0.678 & 0.542 & {$\underline{0.599}$} & 0.758 \\
			$\surd$ & $\surd$ & $\surd$ & {$\mathbf{0.991}$} & {$\mathbf{0.965}$} & {$\mathbf{0.676}$} & {$\mathbf{0.978}$} & {$\mathbf{0.812}$} & {$\mathbf{0.627}$} & {$\mathbf{0.667}$} & {$\underline{0.767}$} \\
\bottomrule
\end{tabular}
\vspace{1ex}
\begin{flushleft}
\end{flushleft}
\label{tab:CTV_comparison}
\end{table*}
	
\subsection{Visualization Analysis}
This section presents qualitative visualizations of the proposed method against four ultrasound-specific nodule detection methods (the methods of Liu \cite{15liu2019automated}, Meng \cite{65Meng2023}, Wang \cite{56wang2023thinking} and Zhou \cite{31zhou2024thyroid}) on a set of randomly selected ultrasound images. As shown in Fig. 8, the comparative methods exhibit over-detection, generally producing lower confidence scores. In contrast, our method achieves more accurate localization and higher detection accuracy for both typical and challenging nodules. These qualitative findings are consistent with the quantitative results and further support the suitability of the proposed method for ultrasound nodule detection, highlighting its promising potential for clinical application.

\begin{figure}[htbp]
    \centering
    \includegraphics[width=0.8\linewidth]{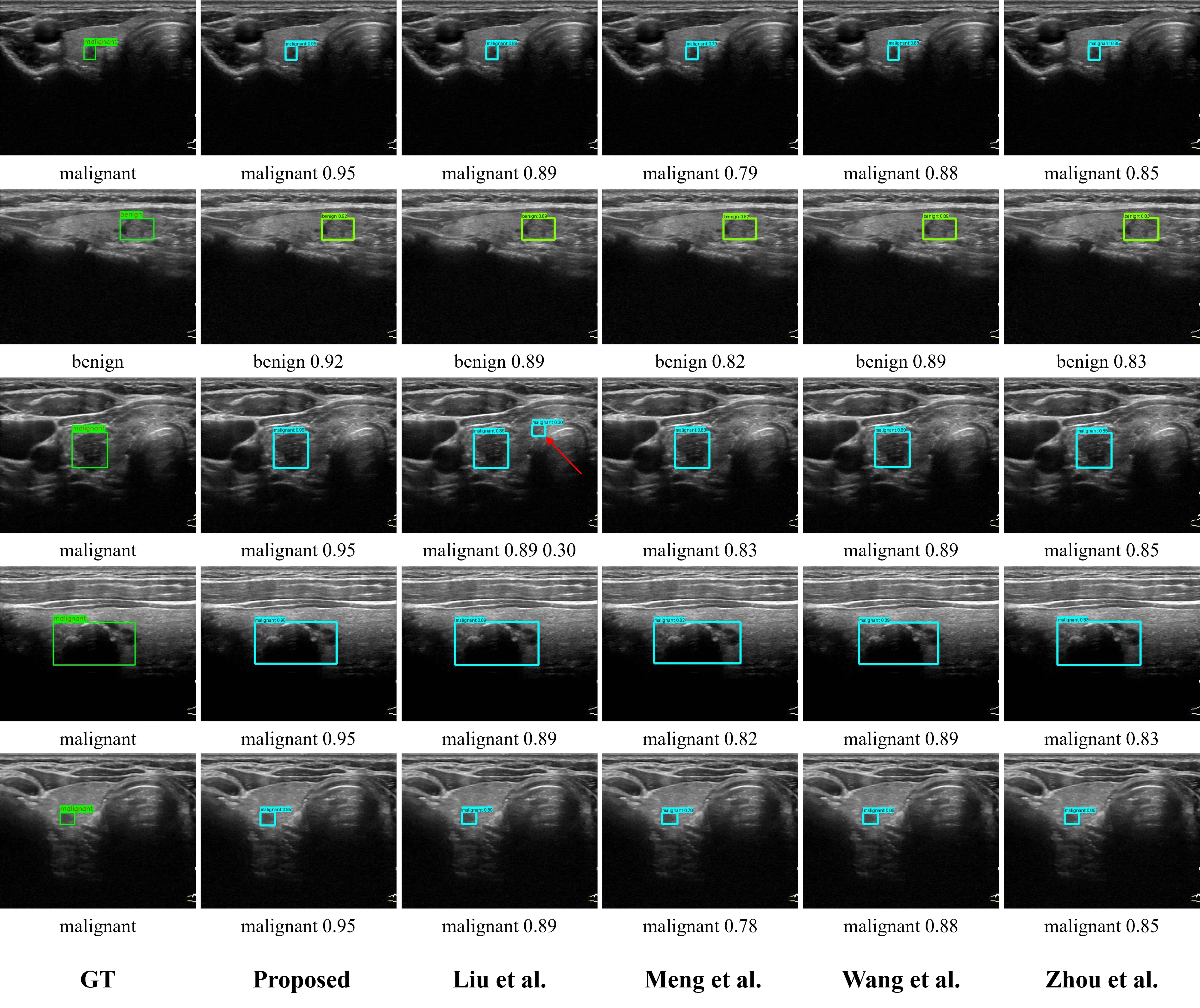} 
    \caption{Visualizations of nodule detection results by different methods on Thyroid I dataset. The red arrow indicates over-detection.} 
    \label{fig:fig8} 
\end{figure}
	
\section{Discussion and Conclusion}
		
In this study, we present a novel prior-guided DETR designed to detect ultrasound nodules. Unlike purely data-driven methods, our method progressively integrates geometric and structural priors to address irregular nodule morphology and speckle noise. By embedding these priors into deformable sampling and spatial-frequency feature extraction, and propagating them via dense feature interaction, the framework ensures consistent multi-level semantic guidance. Extensive experiments across internal, external, and cross-organ datasets have demonstrated that this prior-guided method achieves state-of-the-art performance with superior robustness against ambiguous boundaries and multi-scale variability.
	
In clinical practice, accurate diagnosis is often constrained by the subjective interpretation of noisy ultrasound images. By encoding expert insights into mathematical priors, our framework standardizes this cognitive process, effectively distinguishing anatomical structures from artifacts. This approach offers significant clinical value by mitigating reliance on subjective experience and reducing radiologist workload, thereby reducing inter-observer variability and minimizing missed detections. These findings underscore the potential of explicitly modeling prior knowledge to advance reliable, interpretable computer-aided diagnosis.
	
Future work will extend the framework to 3D ultrasound imaging to leverage volumetric spatial consistency, explore model compression strategies for real-time deployment on portable devices, and conduct prospective clinical validation to further support its translation into routine clinical workflows.

\bibliographystyle{unsrt}
\bibliography{reference}
\end{document}